\documentclass[conference]{IEEEtran}
\usepackage{times}

\usepackage[numbers, sort]{natbib}
\usepackage{multicol}
\usepackage{graphicx}
\usepackage{float}
\usepackage{amsmath}
\usepackage{amssymb}
\usepackage{array}
\usepackage{booktabs}
\usepackage{multicol,multirow}
\usepackage{setspace}
\usepackage{makecell}
\usepackage[dvipsnames]{xcolor}
\usepackage{tasks}
\usepackage{pifont}
\usepackage{colortbl}
\usepackage{xcolor}
\usepackage{framed}
\usepackage{titletoc}
\usepackage{color}
\usepackage{caption, subcaption, overpic, textpos}
\usepackage{titlesec}
\usepackage{comment}

\definecolor{Area_Chair}{RGB}{136,105,160} % DeepPurple
\definecolor{Reviewer_2hnu}{RGB}{59,125,35} % LightGreen
\definecolor{Reviewer_vQ7T}{RGB}{218,165,32} % yellow
\definecolor{Reviewer_F7w2}{RGB}{70,130,180} % LightSlateBlue

\titleformat{\subsubsection}[runin]{\normalfont\itshape}{}{1em}{}[\newline]
\titlespacing{\subsubsection}{0pt}{\baselineskip}{0pt} % 

\usepackage{epigraph}
\newlength\savewidth
\newcommand{\tablestyle}[2]{\setlength{\tabcolsep}{#1}\renewcommand{\arraystretch}{#2}\centering\footnotesize}
\renewcommand{\paragraph}[1]{\vspace{1.25mm}\noindent\textbf{#1}}

\newcolumntype{x}[1]{>{\centering\arraybackslash}p{#1pt}}
\newcolumntype{y}[1]{>{\raggedright\arraybackslash}p{#1pt}}
\newcolumntype{z}[1]{>{\raggedleft\arraybackslash}p{#1pt}}

\newcommand{\app}{\raise.17ex\hbox{$\scriptstyle\sim$}}

\definecolor{deemph}{gray}{0.6}

\definecolor{baselinecolor}{gray}{.9}
\definecolor{yellow}{RGB}{218,165,32}
\definecolor{LightSlateBlue}{RGB}{70,130,180}
\definecolor{DeepBlue}{RGB}{65,100,170}
\definecolor{DeepPurple}{RGB}{136,105,160}
\definecolor{LightGreen}{RGB}{59,125,35}
\definecolor{LightRed}{RGB}{227,120,117}
\newcommand{\baseline}[1]{\cellcolor{baselinecolor}{#1}}
\newcommand{\underfigtab}{\vspace{-10pt}}
\definecolor{cvprblue}{rgb}{0.21,0.49,0.74}
\usepackage[pagebackref,breaklinks,colorlinks,citecolor=cvprblue]{hyperref}

\usepackage[capitalize]{cleveref}
\crefname{section}{Sec.}{Secs.}
\Crefname{section}{Section}{Sections}
\Crefname{table}{Table}{Tables}
\crefname{table}{Tab.}{Tabs.}

\usepackage{xspace}
\newcommand{\modelname}{MPI\xspace}

\begin{document}

\title{
Learning Manipulation by Predicting Interaction
}

% You will get a Paper-ID when submitting a pdf file to the conference system
% \author{Author Names Omitted for Anonymous Review. Paper-ID 
% [6]}

%\author{\authorblockN{Michael Shell}
%\authorblockA{School of Electrical and\\Computer Engineering\\
%Georgia Institute of Technology\\
%Atlanta, Georgia 30332--0250\\
%Email: mshell@ece.gatech.edu}
%\and
%\authorblockA{Twentieth Century Fox\\
%\and
%\authorblockN{James Kirk\\ and Montgomery Scott}
%\authorblockA{Starfleet Academy\\
%San Francisco, California 96678-2391\\

% avoiding spaces at the end of the author lines is not a problem with
% conference papers because we don't use \thanks or \IEEEmembership

\author{\authorblockN{Jia Zeng$^{1,*}$,
Qingwen Bu$^{2,1,*}$,
Bangjun Wang$^{2,1,*}$, 
Wenke Xia$^{3,1,*}$,
Li Chen$^{1}$, 
Hao Dong$^{4}$, \\
Haoming Song$^{1}$,  
Dong Wang$^{1,\natural}$, 
Di Hu$^{3}$,
Ping Luo$^{1}$,
Heming Cui$^{1}$, \\
Bin Zhao$^{1,5,\natural}$,
Xuelong Li$^{1,6}$,
Yu Qiao$^{1}$
and
Hongyang Li$^{1,2,\natural}$
}
\smallskip
\authorblockA{
$^{1}$Shanghai AI Lab
~$^{2}$Shanghai Jiao Tong University
~$^{3}$Renmin University of China \\
$^{4}$Peking University
~$^{5}$Northwestern Polytechnical University 
~$^{6}$TeleAI, China Telecom Corp Ltd
\\
$^{*}$Equal contribution.
$^{\natural}$Corresponding authors.
}
% School of Electrical and Computer Engineering\\
% Georgia Institute of Technology,
% Atlanta, Georgia 30332--0250\\ Email: mshell@ece.gatech.edu}
% \authorblockA{\authorrefmark{2}Shanghai Jiao Tong University}
% \authorblockA{\authorrefmark{3}Starfleet Academy, San Francisco, California 96678-2391\\
% Telephone: (800) 555--1212, Fax: (888) 555--1212}
% \authorblockA{\authorrefmark{4}Tyrell Inc., 
% 123 Replicant Street, Los Angeles, California 90210--4321}
}

\noindent
\twocolumn[{%
\renewcommand\twocolumn[1][]{#1}
\maketitle
\vspace{-5mm}
\begin{center}
    \centering
    \captionsetup{type=figure}
    \includegraphics[width=0.9\textwidth]{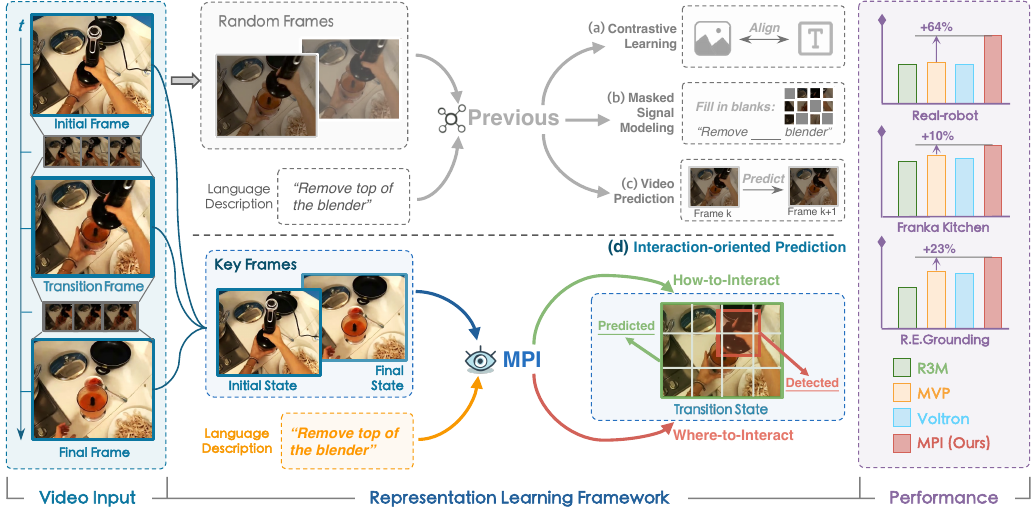}
    \vspace{-6pt}
    \captionof{figure}{{\color{DeepBlue}\textbf{\modelname}} is an interaction-oriented representation learning pipeline for robotic manipulation. 
    Diverging from prior arts grounded in (a) Contrastive Learning, (b) Masked Signal Modeling, or (c) Video Prediction using random frames, our proposed approach in (d) instructs the model towards predicting transition frames and detecting manipulated objects with keyframes as input. 
    As such, the model fosters better comprehension of ``how-to-interact'' and ``where-to-interact''. 
    MPI acquires more informative representations during pre-training and achieves evident improvement across downstream tasks.
    \label{fig:teaser}
    \underfigtab
    }
    \vspace{5mm}
\end{center}%
}]

\begin{abstract}
Representation learning approaches for robotic manipulation have boomed in recent years. 
Due to the scarcity of in-domain robot data, prevailing methodologies tend to leverage large-scale human video datasets to extract generalizable features for visuomotor policy learning. 
Despite the progress achieved, prior endeavors disregard the interactive dynamics that capture behavior patterns and physical interaction during the manipulation process, resulting in an inadequate understanding of the relationship between objects and the environment. 
To this end, we propose a general pre-training pipeline that learns Manipulation by Predicting the Interaction (MPI) and enhances the visual representation.
Given a pair of keyframes representing the initial and final states, along with language instructions, our algorithm predicts the transition frame and detects the interaction object, respectively. 
These two learning objectives achieve superior comprehension towards ``how-to-interact" and ``where-to-interact".
We conduct a comprehensive evaluation of several challenging robotic tasks.
The experimental results demonstrate that \textbf{\modelname} exhibits remarkable improvement by 10\% to 64\% compared with previous state-of-the-art in real-world robot platforms as well as simulation environments. Code and checkpoints are 
publicly shared at \texttt{\url{https://github.com/OpenDriveLab/MPI}}. 
\end{abstract}

\IEEEpeerreviewmaketitle

\section{Introduction}

Visuomotor control of robotic systems entails perceiving and interpreting the surrounding environment from visual inputs, making informed decisions, and executing appropriate actions. 
This capability is of great significance to a wide range of robotic applications, encompassing object manipulation~\cite{brohan2022rt1}, grasping~\cite{zeng2022robotic}, navigation~\cite{shah2023vint, shah2022robotic, NEURIPS2022_286a371d}, \textit{etc}. 
Motivated by the achievements of large-scale pre-training in fundamental vision~\cite{he2022mae, radford2021learning, he2019moco} and natural language processing~\cite{devlin2018bert, sanh2019distilbert}, the field of robotics seeks to utilize large-scale data to build generalizable representations. 
However, for robot manipulation, collecting demonstrations is both laborious and costly. Consequently, the exploration of representation learning methodologies that circumvent dependence on limited in-domain robotics data has emerged as a prominent and trending research focus.

Notably, recent efforts have harnessed large-scale egocentric human video datasets, such as Ego4D~\cite{grauman2022ego4d}, Something-Something V2~\cite{goyal2017something}, and Epic Kitchens~\cite{damen2018epic_kitchens}, to bootstrap representation learning for robot manipulation.
% While previous arts adopt successful approaches in computer vision, incorporating contrastive learning, masked image modelling and video predictive learning (as shown in Fig~\ref{fig:comparison}), to learn generalizable representations for robot manipulation~(visuomotor control), they all overlooked the interactive nature of robot manipulation
As illustrated in Fig.~\ref{fig:teaser}(a)-(b), previous approaches extensively employ contrastive learning~\cite{nair2023r3m, ma2023liv} and masked signal modeling~\cite{radosavovic2023mvp, nair2023r3m, karamcheti2023voltron}. 
While they offer valuable insights for enhancing the performance of robotic systems in downstream tasks, their primary focus tends to be either discerning high-level semantic cues or capturing fine-grained pixel information. 
This focus results in the overlook of crucial interactive dynamics~\cite{billard2019trends, seo2023masked},
which refers to the behavior patterns and physical interactions occurring between a robot and the environment.

In most contexts where robotic systems are deployed, their functions extend beyond passive perceptual capabilities~\cite{ren2015faster} to active engagement with the environment~\cite{brohan2022rt1,brohan2023rt2}.
%thereby instantly responding to changes in perceptual information. 
This prompts an exploration of how to connect interactions that shape the world and visual representations that speak the world. 
One viable approach in this regard is video prediction pre-training~\cite{wu2023gr1}, as depicted in \cref{fig:teaser}(c). Through learning to predict future frames, the model inherently acquires the ability to represent the temporal evolution of scenes~\cite{bartal2024lumiere}. 
However, in the specific context of robot manipulation scenes,
% unlike general scenarios, 
objects typically do not exhibit autonomous movement, and scene changes primarily arise from interaction-driven movements. 
Conventional works~\cite{finn2017deep, gupta2022maskvit} for predicting future consecutive frames merely model the temporal relationships between frames. This task setting is less challenging and prone to introducing noise or redundant information, thus hindering the model's ability to discern interaction-relevant patterns or capture the dynamic interactions in manipulation scenarios effectively~\cite{du2023video}.
Therefore, it is imperative to propose an enhanced interaction-oriented video prediction method.

\begin{figure*}[!t]
    \centering
    \includegraphics[width=\linewidth]{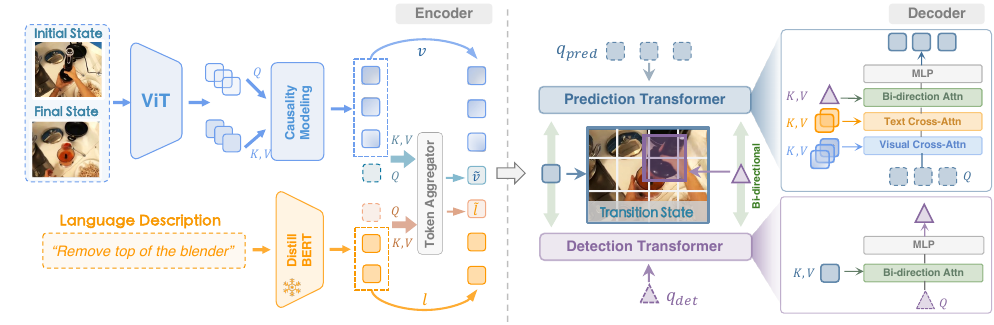}
    % \vspace{-12pt}
    \caption{% \textcolor{blue}{[FIG REVISED]}
    \textbf{The pipeline for pre-training}. \modelname comprises a \textit{multi-modal transformer encoder} and a \textit{transformer decoder} designed for predicting the image of the target interaction state and detecting interaction objects respectively. We achieve synergistic modeling and optimization of the two tasks through information transition between the prediction and detection transformers. The decoder is solely engaged during the pre-training phase while deprecated for downstream adaptations.}
    \label{fig:pipeline}
\end{figure*}

To address the aforementioned challenges, we propose \textbf{\modelname}, which stands for learning \textbf{M}anipulation by \textbf{P}redicting the \textbf{I}nteraction. 
%\modelname is a pre-train framework facilitating the comprehension of manipulation tasks by predicting the interaction process.
%
% which complies with the human tendency to meditate the process before manipulation.
%
As shown in Fig.~\ref{fig:teaser}(d), we formulate two primary objectives, which are also key elements that formulate an interaction, summarized as ``how-to-interact'' and ``where-to-interact''.
Correspondingly, we propose two modules: Prediction Transformer and Detection Transformer.
Given a pair of keyframes representing the initial and final states of an interaction process, along with language instruction, the Prediction Transformer predicts the unseen transition frame that represents the interaction between these two states. 
By employing this objective, the model comprehends ``how-to-interact".
Moreover, the Detection Transformer infers the location of the interaction object in the unseen frame, enabling the model to acquire knowledge of ``where-to-interact''. 
As illustrated in \cref{fig:pipeline}, within a transformer-based encoder-decoder architecture, we leverage a set of prediction queries to estimate the transition frame and a detection query to infer the position of the interaction object.  
The knowledge across the transition frame prediction and interaction object detection is woven through cross-attention among queries, fostering mutual support in the training process. 
As such, 
% Ever since 
we construct a unified 
% and concise 
prediction-detection pre-training framework tailored for robot manipulation. In contrast to predicting future consecutive frames, our algorithm filters out the interaction-irrelevant information and emphasizes the key states, leading to a concentrated learning process. 

To evaluate the effectiveness of \modelname, we assemble a set of downstream tasks: 
1) policy learning on a real-world Franka Emika Panda robot operating in complex scenarios;
2) policy learning on Franka Kitchen~\cite{gupta2020relay} within a complex kitchen environment;
3) manipulation of position-varying objects in Meta-World~\cite{yu2020metaworld} simulation environment; 
and 4) a robotics-related recognition task (\textit{i.e.}, referring expression grounding~\cite{wang2021ocid}), in which the model locates a specified object based on natural language descriptions. 
% %
% %
Extensive experimental results demonstrate the superior performance of our method compared to R3M~\cite{nair2023r3m}, MVP~\cite{radosavovic2023mvp}, and Voltron~\cite{karamcheti2023voltron}. As presented in ~\cref{fig:teaser},
in comparison to MVP, \modelname exhibits a relative improvement rate of 64\% in real-world robot experiments, 10\% on Franka Kitchen, and 23\% (AP@0.75IoU) for Referring Expression Grounding (R.E.G.) task.

\textbf{Contributions.} Our contributions are three folds: 
\textbf{1)} We propose an interaction-oriented representation learning framework, \modelname, towards robot manipulation. By learning interaction, \modelname strengthens the model's comprehension of interactive dynamics in manipulation scenes.
\textbf{2)} 
We introduce the Prediction and Detection Transformer to predict the transition frame and detect the interaction object. These two learning objectives facilitate the model to foster an understanding of ``how-to-interact" and ``where-to-interact".
Moreover, these two learning objectives can be mutually reinforced.
\textbf{3)}
The experimental results reveal that \modelname yields state-of-the-art performance on a broad spectrum of downstream tasks.

\section{Related Work}
\subsection{Representation Learning for Robotics}\label{RelatedWork_RepresentationLearning}

Effective visual representation plays a crucial role in facilitating robots' interaction with the environment~\cite{hansen2022pre, burns2023makes, chen2023end, wu2023policy, yang2023vidar}. Given the limited availability of in-domain robot data, current methods attempt to leverage pre-trained visual models to enhance the generalizability of representations, thereby benefiting downstream manipulation tasks~\cite{nair2023r3m,radosavovic2023mvp}. 
Previous studies have demonstrated the utility of pre-trained models on general vision datasets such as CLIP~\cite{radford2021learning} and ImageNet~\cite{deng2009imagenet} for improving performance on robotic tasks~\cite{shridhar2022cliport,parisi2022unsurprising,khandelwal2022simple}. To further enhance these representations, recent efforts~\cite{radosavovic2023mvp,nair2023r3m} have focused on pre-training visual models using large-scale egocentric datasets~\cite {grauman2022ego4d,goyal2017something}, aiming to bridge the gap between pre-trained representations and downstream robotic scenarios. 
In terms of pre-training techniques, innovative learning approaches have been adopted to abstract representations for robotics. 
For instance, TCN~\cite{sermanet2018time-contrastive} and VIP~\cite{ma2022vip} propose time-contrastive learning to understand action sequences over time, while the recent work MVP~\cite{radosavovic2023mvp} employs masked image modeling method MAE~\cite{he2022mae} for fine-grained object recognition. 
Drawing inspiration from advancements in multi-modal foundation models~\cite{li2023roboflamingo}, recent studies such as R3M~\cite{nair2023r3m}, Voltron~\cite{karamcheti2023voltron}, and LIV~\cite{ma2023liv} bring language into the pre-training 
% stage 
to improve semantic comprehension for robotic tasks.

Recently, there 
% have been 
are 
interesting 
% findings about
attempts on the practical effectiveness of representation learning using ego-centric human data~\cite{dasari2023unbiased, burns2023Robust_Representations}.
Burns \textit{et al.}~\cite{burns2023Robust_Representations} claim that
models pre-trained on manipulation-relevant data do not generalize better than those trained on standard pre-training datasets.  
Based on the MAE pre-train paradigm, Dasari \textit{et al.}~\cite{dasari2023unbiased} demonstrate that representations derived from traditional computer vision datasets such as ImageNet outperform those derived from Ego4D. 
However, we argue that the experimental phenomena arise because previous literature solely applies existing pre-training techniques to large-scale egocentric datasets, without explicit consideration of the interaction characteristics during manipulation processes. As such, they fail to fully utilize the advantages of 
% representing 
modeling temporal relationships and interaction in manipulation-related datasets.
In this work, we incorporate interaction-oriented learning to embed representations with manipulation task-specific knowledge, aiming to enhance performance in downstream robotic applications.

\subsection{Learning Interaction in Robotics}

Gaining a comprehensive understanding of interaction is essential for effectively manipulating objects. Past research on learning interaction can be broadly classified into two main categories: explicit representation and implicit encoding.

\paragraph{Explicit Representation.} 
Within the field of robotics, affordance~\cite{gibson1978ecological} refers to the properties of an object that enable specific interactions, \textit{i.e.}, a perceptual cue for object manipulation. Previous works have explicitly expressed interaction information by estimating affordance, which can take various forms, including contact points~\cite{li2023manipllm}, grasp point pairs~\cite{zeng2022robotic}, or heatmaps~\cite{li2023locate, mees2023grounding, chen2023affordance}.
By leveraging internet videos of human behavior, \citet{bahl2023affordances} train a visual affordance model capable of predicting contact point and trajectory waypoints, which explicitly represent the locations and manners in which humans interact within a scene. Building upon such affordances, SWIM~\cite{mendonca2023structured} constructs affordance-space world models that enable robots to learn manipulation skills. 
%
% Although explicit representations are more interpretable, implicit representations are more flexible.
%
In our proposed pipeline, interaction object detection can be viewed as an explicit representation utilized for supervision. 

\paragraph{Implicit Encoding.} 
Given the increasing prevalence of video generation and synthesis, recent research has investigated the potential of applying advanced diffusion techniques to robot manipulation tasks, aiming to model the interaction process implicitly. Approaches such as UniPi~\cite{du2023learning} leverage internet data and treat the sequential manipulation decision-making problem as a text-conditioned video generation problem. They train a separate inverse dynamics module to estimate actions. Additionally, UniSim~\cite{yang2023unisim} aims to develop a universal simulator for simulating the interaction process and training various policies through generative modeling. 
Instead, our method focuses solely on utilizing the image reconstruction task to guide the representation to incorporate interactive dynamics.

% \section{MPI: Interaction-oriented Learning}
\section{MPI: Manipulation by Predicting Interaction}
\label{sec:method_mpi}

The overall pipeline of our approach is depicted in \cref{fig:pipeline}.
We begin with a triplet of keyframes that represent the initial, transition, and final states in an interaction process. Two of these frames are utilized as inputs for the visual encoder, while the remaining frame acts as the prediction target. To establish the state transition relationship between the input frames, we incorporate a causality modeling module that facilitates dynamic attention between the two states. Furthermore, a language description that corresponds to the entire interaction process is employed to provide high-level semantic cues, subsequently conditioning the decoding process for the target state. A shared token aggregator is devised to generate a concise embedding vector by combining the encoded vision and language representations.

In terms of the decoder, we configure the Prediction Transformer and Detection Transformer for interaction frame prediction and interaction object detection. The information exchange between these two modules allows the model to establish potential relationships between ``how-to-interact'' and ``where-to-interact'', thereby optimizing both objectives.

%\subsection{Core Module Design Specification}

\subsection{Data: A Triplet of Keyframes}
\label{sec:Triplet_Key_Frames}

% \noindent
Based on the annotations provided in the Ego4D Hand-Object Interaction dataset~\cite{grauman2022ego4d}, we define three keyframes denoted as $\{F_{\text{init}}, F_{\text{trans}}, F_{\text{final}}\} \in \mathbb{R}^{3 \times H \times W}$. These frames represent the initial, transition, and final states of an interaction process. The initial state serves as the starting point and establishes the contextual foundation for the interaction process. Therefore, $F_{\text{init}}$ is consistently used as the input during pre-training. The selection between $F_{\text{trans}}$ and $F_{\text{final}}$ is subject to a Bernoulli distribution parameterized by $p \in [0,1]$:
\smallskip
\begin{align}
\label{eq:input}
    F_{\text{input}}  &=
        \begin{cases}
                [F_{\text{init}}, F_{\text{trans}}],  \qquad &\text{if}\  \alpha = 0, \\
                [F_{\text{init}}, F_{\text{final}}],  \qquad &\text{if}\  \alpha = 1, \\
        \end{cases} 
    % \text{where}  & \hspace{0.5em} \alpha \sim \text{Bernoulli}(p) \nonumber.
\end{align}
where $\alpha \sim \text{Bernoulli}(p)$. 
Predicting the final state conditioned on the given motion endows
our model with the ability to estimate consequences.
On the other hand, predicting a transition state enhances the understanding of the causality involved in state transitions.
By learning these two situations simultaneously during training, the model fosters a holistic comprehension of ``how-to-interact" and ``where-to-interact".

\subsection{Encoder: Representation of Interactive Dynamics}
\paragraph{Decoupled Multi-modal Encoder.}
\noindent
The multi-modal encoder plays a central role in learning representations, as illustrated by the components highlighted in {\color{LightSlateBlue}light blue} and {\color{yellow}orange} in \cref{fig:pipeline}.
For our visual encoder design, we adhere to the established Vision Transformer (ViT) framework~\cite{dosovitskiy2020image}.
The input frame $F_{i}$ is divided into non-overlapping patches with a window size $s$, and then transformed into a sequence of visual embeddings $v \in \mathbb{R}^{L_{\text{vis}} \times d}$ suitable for processing by ViT. Here, $L_{\text{vis}}$ represents the length of the visual embedding, and $d$ stands for feature dimension. The value of $L_{\text{vis}}$ is equal to the number of divided windows $HW / s^2$, where $H$ and $W$ denote the height and width of the input frame accordingly.

In line with Voltron~\cite{karamcheti2023voltron}, we adopt DistillBERT~\cite{sanh2019distilbert} as the language encoder. The language description of the interaction process is initially tokenized into a sequence of numerical identifiers based on their position within a predefined vocabulary set $V$, and then padded to the maximum length $L_{\text{lang}}$. Consistent with previous multi-modal representation learning frameworks~\cite{karamcheti2023voltron,nair2023r3m}, the language encoder remains frozen (without parameter updates). In addition, our visual encoder and language encoder are decoupled, enabling the omission of language input in downstream tasks.

%\paragraph{Encoder: Causality Modeling}

% \noindent
\paragraph{Causality Modeling.} The causality modeling module takes as input the visual embedding $\{v_{0}, v_{1}\} \in \mathbb{R}^{L_{\text{vis}} \times d}$ from two parallelly encoded frames. The embedding corresponding to the initial state is used as queries,  while the subsequent state serves as both keys and values. These encoded frames undergo processing through a deformable attention layer~\cite{zhu2021deformable} to capture patch-to-patch relations. 
% The initialization of reference points for feature sampling within the deformable attention mechanism aligns with the partitioning of the original image into patches. Learnable parameters regulate the sampling offset, thereby enabling adaptive causality modeling between two input frames. 
This process can be elucidated as:
% \smallskip
\begin{equation}
\begin{aligned}
  v'_{0} & = \texttt{DeformAttn}(Q = v_{0}, K = V = v_{1}),\\
  v \ & = \texttt{Norm}(v_{0} + v'_{0}).
\end{aligned}
\end{equation}
% \smallskip
The merged visual representation $v \in \mathbb{R}^{L_{\text{vis}} \times d}$, which encodes causality relationships between the input states, is then leveraged in the decoder. In addition, this module reduces the computational burden in subsequent stages by combining the embeddings of the two frames into a unified representation. For downstream adaptations with single-image input, we simply replicate image features from the ViT to generate inputs for this module.

%\paragraph{Encoder: Token Aggregator}
%\label{sec:aggregator}

% \noindent
\paragraph{Token Aggregator.}
\label{sec:aggregator} 
We introduce the token aggregator, which is motivated by two primary considerations: \textbf{1)} To address the challenge posed by disparate lengths between visual and language embeddings $\{v, l\}$, we aim to enable vision encoders to generate language-aligned, semantic-rich features by minimizing the distance between aggregated tokens $\{\Tilde{v}, \Tilde{l}\}$, and \textbf{2)} By employing aggregated tokens as concise representations for downstream tasks, we facilitate fair comparisons with prior methods~\cite{nair2023r3m,radosavovic2023mvp} that inherently rely on aggregated features.

As discussed in Voltron~\cite{karamcheti2023voltron}, multiheaded attention pooling (MAP)~\cite{lee2019set} proves to be significantly more effective compared to alternative prevalent approaches, such as the \texttt{[cls]} token~\cite{radosavovic2023mvp} or global pooling~\cite{nair2023r3m}, for deriving compact representations from a sequence of feature embeddings. 
Following this insight, we also employ a MAP block, shared between vision and language representations, to extract aggregated tokens. The aggregation process begins with employing zero-initialized latent embedding vectors, denoted as $\{\Tilde{v}, \Tilde{l}\} \in \mathbb{R}^{d}$ for vision and language, respectively. Taking the visual part as an example, this process can be formulated as follows: 

\begin{equation}
\begin{aligned}
   \Tilde{v} & = \texttt{Norm}(\Tilde{v} + \texttt{Attn}( Q = \Tilde{v}, K=V= v)), \\
    \Tilde{v} & = \texttt{Norm}(\Tilde{v} + \texttt{MLP}(\Tilde{v})). 
\end{aligned}
\end{equation} 

\subsection{Decoder: Prediction and Detection}
%\paragraph{Decoder: Prediction Former}

% \noindent
\paragraph{Prediction Transformer.} The detailed structure of the Prediction Transformer is shown in {\color{DeepBlue}deep blue} on the right side of \cref{fig:pipeline}.
Given visual and language embeddings from the encoder, the Prediction Transformer is responsible for predicting pixels of the frame that represent the unseen interaction state. In contrast to the original MAE~\cite{he2022mae} where queries are used to represent masked regions in the input image, our prediction process starts with a set of randomly initialized queries $q_\text{pred} \in \mathbb{R}^{L_{\text{vis}} \times d}$, which correspond to each patchified window of the target frame. To obtain the necessary information for accurate prediction, we perform cross-attention between the visual and language embeddings $\{v, l\}$ and use $q_{\text{pred}}$ as queries. Additionally, we enhance the prediction queries by incorporating semantic information related to the interaction object, as encapsulated in the detection query $q_{\text{det}}$, using a bidirectional attention mechanism.
After the iterative processing within the multi-layer decoder, the prediction queries are linearly projected (with $\mathcal{F}$) to generate pixel-level prediction results denoted as $\hat{F}{_\text{pred}} = \mathcal{F}(q{_\text{pred}}) \in \mathbb{R}^{3 \times H \times W}$.

%\paragraph{Decoder: Detection Former}

% \noindent
\paragraph{Detection Transformer.} Taking into consideration the object-centric nature of an interaction process~\cite{devin2018deep}, we introduce a detection query $q_{\text{det}} \in \mathbb{R}^{d}$ to locate the interaction object, as highlighted in {\color{DeepPurple}purple} in \cref{fig:pipeline}. To expedite training convergence, we initialize the detection query with the aggregated visual embedding $\Tilde{v}$. It is noteworthy that the detection query is designed to regress the object location mainly based on the information inferred from the prediction queries $q_{pred}$, which we expect to contain comprehensive information about the target state. The exchange of information occurs through bidirectional attention between the two sets of queries. A subsequent two-layer multi-layer perceptron (MLP) is applied to obtain the regression box: $\hat{b}_{\text{det}} = \text{MLP}(q_{\text{det}}) \in \mathbb{R}^{4}$.

Our empirical analysis reveals that, despite the slow convergence observed in the early training phases due to inadequately informative representations on both ends, the transmission of query information facilitates eventual convergence for both tasks. The integrated modeling and optimization of these two tasks synergistically enhance the model's capacity to acquire a more effective representation for robot manipulations.

\subsection{Optimization Objectives}

The primary objective of our proposed framework during pre-training is to predict the unseen state and detect interaction objects, which are framed as ``how-to-interact" and ``where-to-interact", respectively. For target frame prediction, we utilize mean-squared error as the loss function, while for detection, we supervise the model with L1 loss and GIoU loss~\cite{rezatofighi2019generalized}. Furthermore, our preliminary experiments demonstrate that incorporating contrastive loss $\mathcal{L}_{\text{con}}$ between aggregated visual and language embeddings (\textit{i.e.}, $\Tilde{v}$ and $\Tilde{l}$) improves representation learning. In summary, our pre-training framework aims to minimize the following components of loss:

\begin{equation}
\label{eq:loss}
\begin{aligned}
    &\mathcal{L}_{\text{pred}} = \text{MSE} ( \hat{F}_{\text{pred}}, F_{\text{gt}} ), \\
    &\mathcal{L}_{\text{det}} \ = \text{L1}(\hat{b}_{\text{det}}, b_{\text{gt}}) + \text{GIoU}(\hat{b}_{\text{det}}, b_{\text{gt}}), \\
    &\mathcal{L}_{\text{con}} \,= \text{InfoNCE} (\Tilde{v}, \Tilde{l}), \\
    &\mathcal{L} \ \ \ \,= \lambda_{1}\mathcal{L}_{\text{pred}} + \lambda_{2}\mathcal{L}_{\text{det}} + \lambda_{3}\mathcal{L}_{\text{con}},
\end{aligned}
\end{equation}
where $F_{\text{gt}}$ represents the target frame, $b_{gt}$ denotes the bounding box annotation, and $\lambda_{1}, \lambda_{2}, \lambda_{3}$ are balancing factors for the different learning objectives. The InfoNCE loss~\cite{oord2018representation} employs cosine similarity as the distance measure.

\begin{figure*}
    \centering
    \includegraphics[width=\linewidth]{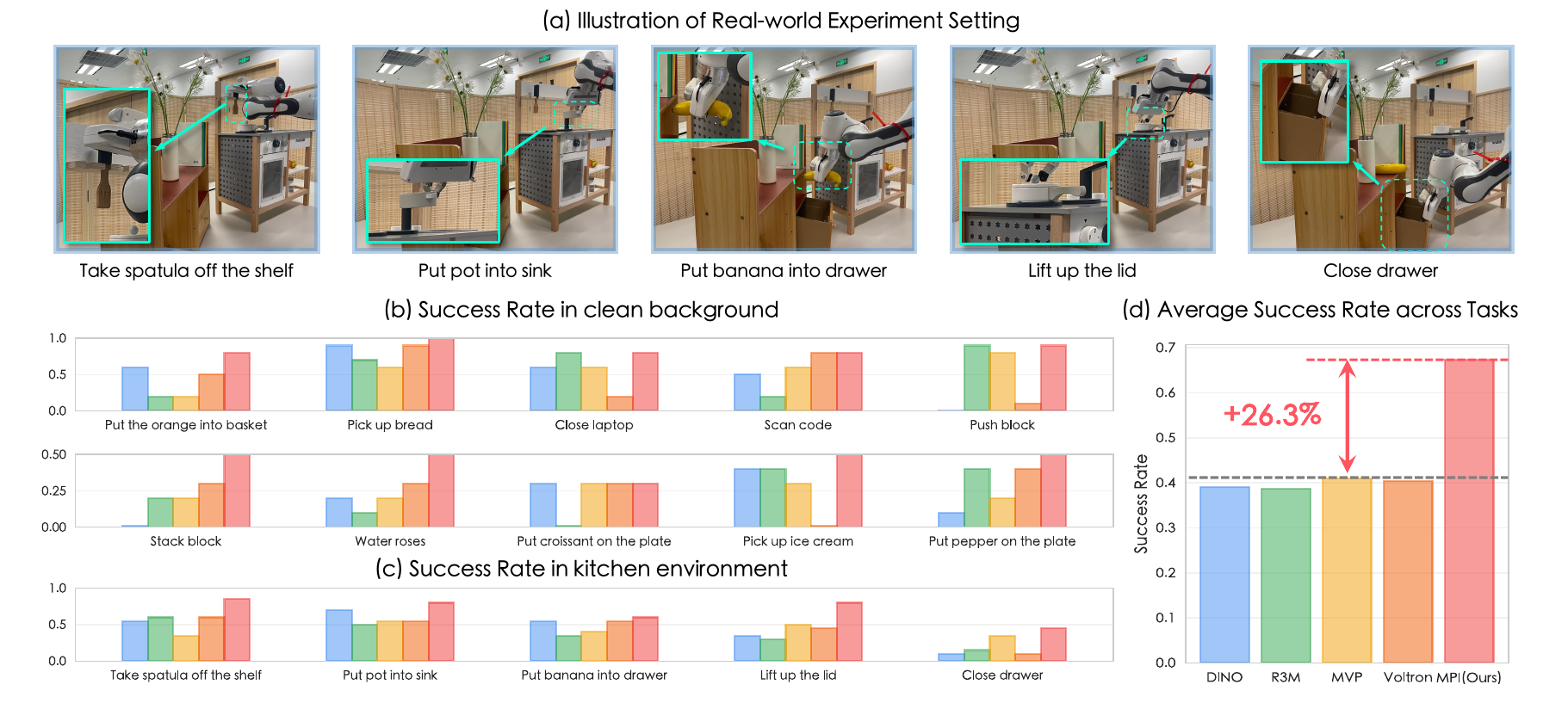}
    \vspace{-20pt}
    \caption{\textbf{Real-world robot experiments.} (a) Illustrations of real-world experiments in the kitchen environment. (b) Detailed success rate of ten tasks within a clean background. (c) Results of five tasks in the complex kitchen environment. (d) \modelname 
    % remarkably 
    outperforms previous state-of-the-art
    % in our extensive real-world experiments 
    with an average elevation of 26.3\% success rate across 15 tasks.}
    \label{fig:real-world}
    \vspace{-5pt}
\end{figure*}

\section{Experiments}

% Our evaluation suites aim to study the following questions:
% \begin{itemize}
%   \item [Q1.] 
%   Can \modelname employ its ``where-to-interact'' and ``how-to-interact'' capabilities to both real-world and complex simulation applications?      
%   \item [Q2.]
%   Can \modelname boast stronger ``where-to-interact'' performance on language-conditioned visual grounding?
%   \item [Q3.]
%   Can \modelname demonstrate its ``how-to-interact''  advantages in interaction-oriented simulation tasks?
% \end{itemize}

% \begin{enumerate}
%     \item Q1. 
%     \item Q2. 
%     \item Q3. 
% \end{enumerate}

\subsection{Evaluation Suites}
\label{sec:Evaluation_Suites}
We have devised a comprehensive evaluation suite consisting of four robot learning tasks. Specifically, the suite comprises the following components: 
\begin{enumerate}
    \item Visuomotor control on real-world robots~(\cref{sec:exp_real_robot}): This task aims to validate the applicability of our proposed method in various tasks in real-world scenarios, and assess the generalization capabilities.
    \item Visuomotor control on Franka Kitchen~(\cref{sec:exp_franka}): In this task, we benchmark the performance of our approach in complex simulation environments, comparing it against the previous state-of-the-art.
    \item Visuomotor control on Meta-World~(\cref{sec:exp_metaworld}): The target is to verify the ability to identify interaction objects with unfixed locations and perform manipulation.
    \item Referring expression grounding~(\cref{sec:exp_REG}): This task evaluates the representation of object-centric semantics and spatial relationships conditioned by language.
\end{enumerate}

Following the established practice in previous literature~\citep{radosavovic2023mvp,nair2023r3m,karamcheti2023voltron}, each evaluation includes adaptation data and evaluation metrics. 
% The adaptation data consists of both visual~(as RGB frames) and language~(an instruction for imitation learning) input. 
We evaluate \modelname and various baseline models by freezing the pre-trained vision encoders and adapting task-specific ``heads'' on top of the extracted representations. In the following sections, we provide the rationale for selecting each evaluation task and present the corresponding experimental results. Furthermore, we delve into ablations on pre-training paradigms and model designs in~\cref{sec:exp_ablation}.

\subsection{Pre-training Details}

Due to the labor-intensive nature of collecting robot manipulation data, we leverage large-scale egocentric human video datasets as the source for pre-training. 
Instead of using the entire Ego4D dataset~\cite{grauman2022ego4d}, we utilize the hand-object interaction subset only. This dataset contains recordings of dynamic interaction processes during manual manipulation, captured with head-mounted cameras.
Each video clip is accompanied by a textual description of the ongoing action.
Additionally, the dataset annotates keyframes that capture critical moments, including the pre-change, change onset, and post-change states, which we refer to as the initial, transition, and final states in our paper.
The bounding box annotations for the interaction objects in these three frames are also provided.
We select 93K video clips for pre-training.

All models are pre-trained on 8 NVIDIA A100 GPUs. The ViT-Small version is trained for 200 epochs, using an initial learning rate of 1e-4 and a batch size of 64 per GPU. We choose the AdamW~\cite{loshchilov2017decoupled} optimizer. Training ViT-Small versions take approximately 15 hours. The ViT-Base version follows most of the same settings as the small version but is trained for a total of 400 epochs. 
In our experiments, we assign uniform weights for the loss terms, specifically, $\lambda_{1}$, $\lambda_{2}$ and $\lambda_{3}$ equal to 1, without
adjustments for performance improvement.
Further implementation details can be found in the Appendix, and we release the codes and models publicly for reproduction purposes.

\subsection{Evaluations on Visuomotor Control}

\subsubsection{1) Real-world Robots}
\label{sec:exp_real_robot}

\paragraph{Motivation.} In order to authentically assess the effectiveness of various visual representation learning approaches for enabling efficient robotic learning in real-world environments, we have introduced a series of manipulation tasks.

\paragraph{Evaluation Details.} In the deployment of a Franka Emika Panda robot for a series of tasks, we employ a 3D SpaceMouse to collect 10 teleoperated demonstrations for each task. An Operational Space Controller operating at 20Hz, as detailed in Deoxys~\cite{zhu2022viola}, is utilized. In the evaluation phase, we use the frozen visual encoder to extract visual representations. These visual representations are then concatenated with the robot's proprioceptive state as inputs to a shallow MLP policy head. The objective is to predict continuous delta end-effector poses through action chunking~\cite{zhao2023learning}.
%
% Although language serves as guidance in our pre-training process, the visual encoder and language encoder are decoupled, allowing us to exclude language input in downstream tasks.
%
To ensure fair comparisons with prior works~\cite{nair2023r3m,radosavovic2023mvp}, we solely employ the aggregated visual embedding without any language interference. 
Each task is conducted 20 times, and we report the average success rate.

To provide a comprehensive evaluation of the effectiveness of different pre-trained encoders, we design two distinct scenarios with varying levels of complexity.
The first scenario consists of ten diverse manipulation tasks in a \textit{clean background}. These tasks include 1) putting the orange into the basket, 2) stacking the block, 3) picking up the ice cream, 4) closing the laptop, 5) scanning code, 6) watering roses, 7) putting the croissant on the plate, 8) picking up the bread, 9) pushing the block, and 10) putting the pepper on the plate. These tasks require fundamental manipulation skills such as Pick \& Place, articulated object manipulation, \textit{etc}.

In addition, we construct a more challenging \textit{kitchen environment} that incorporates various interfering objects and backgrounds relevant to the target tasks. In this environment, we present five tasks: 1) taking the spatula off the shelf, 2) putting the pot into the sink, 3) putting the banana into the drawer, 4) lifting the lid, and 5) closing the drawer. As shown in \cref{fig:real-world}(a), the complexity of these scenarios necessitates the visual encoder to possess both the ``how-to-interact'' and ``where-to-interact'' abilities to effectively handle these tasks.

\begin{figure}[t!]
    \centering
    \includegraphics[width=.95\linewidth]{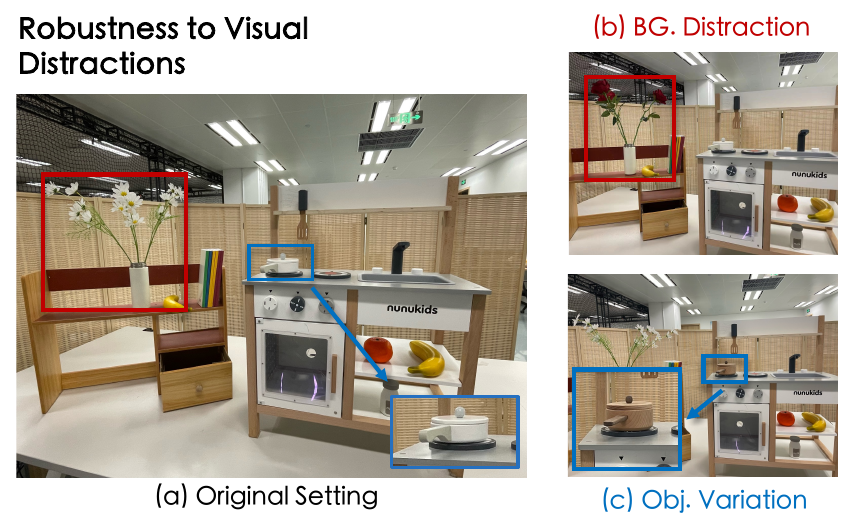}
    \vspace{-4pt}
    \caption{\textbf{Illustration of real-world validation on generalization} to
    % , including 
% evaluation for 
(b) background (BG.) distraction when we put a banana into the drawer, and (c) object variation when we lift the lid.
    }
    \label{fig:distraction}
    \vspace{-2pt}
\end{figure}

\paragraph{Experimental Results.} The results for tasks conducted in a \textit{clean background} are presented in \cref{fig:real-world}(b). While certain models exhibit superior performance in specific tasks (\textit{e.g.}, the Voltron excels in the ``scan code'' task), they fail to consistently maintain high performance across all tasks. In contrast, MPI directs its focus towards the interactive objects and attains a more broadly applicable representation, consequently yielding consistently elevated performance across a spectrum of tasks.

Within the more challenging \textit{kitchen environment}, as shown in \cref{fig:real-world}(c), MPI exhibits outstanding performance across all manipulation tasks. Notably, it demonstrates particular proficiency in tasks that require precise object perception to interact with small objects, such as the ``lift the lid'' and the ``take spatula off''. 
% This enhanced performance is a direct consequence of our interaction-oriented training approach, which substantially improves the encoder's capacity to accurately handle manipulation tasks.
Overall, as demonstrated in \cref{fig:real-world}(d), MPI surpasses prior approaches, leading to an average success rate increase of \textbf{26.3\%} across a broad spectrum of 15 tasks.

\begin{table}[tb!]
  \small
  \centering
  \caption{\textbf{Evaluation on generalization to background distraction and object variation.} 
  % The relative performance degradation due to changes in background or object is listed in the parenthesis. 
  \modelname  performs  best in terms of success rate (\%)
  under interferences nonetheless, benefiting from the robust representations it learns.}
  \label{table:distraction}
  \vspace{-2pt}
  \setlength{\tabcolsep}{1.5mm}{
  \begin{tabular}{l| c c | c c}
    \toprule
         \multirow{2}{*}{Method} & \multicolumn{2}{c|}{Put banana into drawer} & \multicolumn{2}{c}{Lift up the lid}\\
         % \cmidrule{2-3}
         % \cmidrule{4-5}
         & Original & Distraction & Original & Variation  \\
         \midrule
        DINO~\cite{caron2021emerging}& 55 & 35 ($\downarrow$36.4\%) & 35 & 20 ($\downarrow$42.9\%)\\
        R3M~\cite{nair2023r3m} & 35 & 25 ($\downarrow$28.6\%) & 30 & 15 ($\downarrow$50.0\%) \\
        MVP~\cite{radosavovic2023mvp} & 40 & 25 ($\downarrow$37.5\%) & 30 & 15 ($\downarrow$50.0\%)\\
        Voltron~\cite{karamcheti2023voltron} & 55 & 25 ($\downarrow$54.5\%) & 45 &  20 ($\downarrow$55.6\%)\\
        \midrule
        \baseline{\modelname (Ours)} &\baseline{\bf{60}} &\baseline{\textbf{55} ($\downarrow$8.3\%)} &\baseline{\textbf{80}} & \baseline{\textbf{50} ($\downarrow$37.5\%)}\\
    \bottomrule
    \end{tabular}}
    \vspace{-2pt}
\end{table}

\begin{table}[tb!]
  \small
  \centering
  \caption{
  \textbf{Evaluation on generalization to object position and lighting condition.} 
  \modelname performs best in terms of success rate (\%) under various interferences.
  $^{*}$We incorporate variations in lighting along with shifts in position.
  }
  \label{table:generalization_position_lighting}
  \vspace{-2pt}
  \setlength{\tabcolsep}{3.5mm}{
  \begin{tabular}{l| c c }
    \toprule
         {Method} & {Object Position} & {$^{*}$Lighting Condition}\\
         \midrule
        Voltron~\cite{karamcheti2023voltron}  & 30 &  0\\
        MVP~\cite{radosavovic2023mvp} & 40  & 20 \\
        \midrule
        \baseline{\modelname (Ours)} &\baseline{\bf{60}} & \baseline{\bf{30}}\\
    \bottomrule
    \end{tabular}
}
    \vspace{-2pt}
\end{table}

 \begin{table*}[t!]
  \small
  \centering
  \caption{\textbf{Results of single-task visuomotor control on Franka Kitchen.}
  We report the success rate~(\%) over 50 randomly sampled trajectories. We \textbf{bold} the best result for models with similar parameters and \underline{underline} the second. ``INSUP.'' represents classification-based supervised learning on ImageNet. MPI consistently exhibits superior performance across multiple
  % a spectrum of 
  tasks.
  % The visual backbone network of each model, along with its parameters, is also listed.
  %\dag: We adopt our proposed Proprio-conditioned MAP to extract representations for policy learning. For methods based on ResNet50, we retrieve the feature before global average pooling and flatten it to align with the input dimension of MAP.
  }
  \label{table:franka_kitchen}
    \centering
    \begin{tabular}{l | cc | ccccc | c}
    \toprule
    Method &  Backbone & Param. & Turn knob & Open door & Flip switch & Open microwave & Slide door & Average\\
    % \cmidrule{1-9}
    % \midrule
    \midrule
    % \multicolumn{9}{c}{Proprio-conditioned Policy}\\
    % \midrule
    INSUP.~\cite{he2016resnet} & ResNet50 & 25.6M & 28.0 & 18.0 & 50.0 & 26.7 & 75.7 & 39.7\\
    CLIP~\cite{radford2021learning} & ResNet50 & 25.6M & 26.3 & 13.0 & 41.7 & 24.7 & 86.3 & 38.4\\
    R3M~\cite{nair2023r3m}    & ResNet50 & 25.6M & 53.3 & \textbf{50.7} & 86.3 & \underline{59.3} & 97.7 & 69.5\\
    % R3M~\cite{nair2023r3m}\dag   & ResNet50 & 25.6M & 66.3 & 43.7 & 87.3 & 36.3 & \underline{99.7} & 66.7\\
    %\baseline{Ours}        & \baseline{ResNet50} & \baseline{} & \baseline{} & \baseline{} & \baseline{} & \baseline{} & \baseline{} \\
    %\cmidrule{1-9}
    %Voltron~\cite{karamcheti2023voltron}  & ViT-Small & 22M & 16.7 & 18.0 & 29.3 & 12.3 & 82.0 & 31.7 \\
    Voltron~\cite{karamcheti2023voltron}  & ViT-Small & 22M & \underline{71.7} & 45.3 & \textbf{95.3} & 40.3 & \underline{99.7} & \underline{70.5}\\
    % \baseline{Ours} & \baseline{ViT-Small} & \baseline{22M} &  \baseline{\underline{74.0}} & \baseline{43.0} & \baseline{88.3} & \baseline{47.7} & \baseline{\textbf{100.0}} & \baseline{\underline{70.6}} \\
    \baseline{\modelname (Ours)}                & \baseline{ViT-Small} & \baseline{22M} &  \baseline{\textbf{83.3}} & \baseline{\underline{50.3}} & \baseline{\underline{89.0}} & \baseline{\textbf{59.7}} & \baseline{\textbf{100.0}} & \baseline{\bf{76.5}} \\
    % \cmidrule{1-9}
    % \midrule
    \midrule
    %EVA-02~\cite{fang2023eva02}   & ViT-Base & 86M & 26 & 24 & 70 & 30 & 80 & 46.0\\
    %EVA-02~\cite{fang2023eva02}  & ViT-Base & 86M & 75.7 & 47.7 & \underline{92.0} & \underline{42.7} & \textbf{100.0} & 71.6\\/
    %MVP~\cite{radosavovic2023mvp}    & ViT-Base & 86M & 28.3 & 27.3 & 48.3 & 22.0 & 84.3 &42.1\\
    MVP~\cite{radosavovic2023mvp}   & ViT-Base & 86M & \underline{79.0} & \underline{48.0} & 90.7 & \underline{41.0} & \textbf{100.0} & \underline{71.7}\\
    %Voltron~\cite{karamcheti2023voltron} & ViT-Base & 86M & 16.7 & 16.0 & 32.3 & 14.3 & 79.7 & 31.8\\
    Voltron~\cite{karamcheti2023voltron}  & ViT-Base & 86M  & 76.0 & 45.3 & \underline{91.0} & \underline{41.0} & \underline{99.3} & 70.5\\
    % \baseline{Ours}           & \baseline{ViT-Base}  & \baseline{86M} & \baseline{\underline{80.3}} & \baseline{42.0} & \baseline{88.0} & \baseline{\textbf{54.3}} & \baseline{\textbf{100.0}} & \baseline{\underline{72.9}} \\
    \baseline{\modelname (Ours)}               & \baseline{ViT-Base}  & 
     \baseline{86M} & 
     \baseline{\textbf{89.0}} & \baseline{\textbf{57.7}} & \baseline{\textbf{93.7}} & \baseline{\textbf{54.0}} & \baseline{\textbf{100.0}} & \baseline{\textbf{78.9}}     \\     
    \bottomrule
    \end{tabular}
    % \captionof{table}{\textbf{Results of Single-Task Visuomotor Control on Franka Kitchen Simulation Environment.}}
%   We report the success rate~(\%) over 50 randomly sampled trajectories. We \textbf{bold} the best result and \underline{underline} the second.
%   %\dag: We adopt our proposed Proprio-conditioned MAP to extract representations for policy learning. For methods based on ResNet50, we retrieve the feature before global average pooling and flatten it to align with the input dimension of MAP.}
\end{table*}

\begin{table}[tb!]
  \small
  \centering
  \caption{
  \textbf{Robustness evaluation under Franka Kitchen environment}, with varying background distraction and lighting conditions~\cite{burns2023Robust_Representations}. 
  \modelname shows greater robustness against background distraction and lighting condition variation.
  }
  \label{table:franka_robustness_eval}
  \vspace{-2pt}
  \setlength{\tabcolsep}{3.5mm}{
  \begin{tabular}{l| c c | c}
    \toprule
         {Method} & {Distractors} & {Lighting} & {Average}\\
         \midrule
        R3M~\cite{nair2023r3m} & 20.1 &  0 &  10.1\\
        Voltron~\cite{karamcheti2023voltron} & 49.3 &  27.6 &  38.5\\
        MVP~\cite{radosavovic2023mvp} & 51.3 & 24.0 &  37.7 \\
        DINO~\cite{caron2021emerging} & 49.7 & 26.4 &  38.1 \\
        \midrule
        \baseline{\modelname (Ours)} &\baseline{\bf{54.3}} & \baseline{\textbf{45.2}} & \baseline{\textbf{49.7}}\\
    \bottomrule
    \end{tabular}
}
    \vspace{-2pt}
\end{table}

\paragraph{Generalization to background distraction and object variation.}
To assess the generalization capability and robustness of visual representations, we design two distinct scenarios within the complex kitchen environment. The first task involves placing a banana into a drawer. We introduce background distractions by replacing daisies with roses, as illustrated in \cref{fig:distraction}(b). 
In \cref{fig:distraction}(c), 
% to examine the generalization capabilities of various models in response to alterations in the manipulated object, 
we conduct another task: lifting up the pot lid, where we substitute the white pot with a wooden one.

The results are in \Cref{table:distraction}. It is evident that all existing approaches suffer degraded performance when encountering out-of-distribution situations. However, MPI demonstrates remarkable robustness against such disturbances, experiencing merely a modest 8.3\% decrease in the presence of background distraction and a minimum drop of 37.5\% in object variation. These results validate the generalizability of our method to unseen environments and manipulation objects.

\paragraph{Generalization to object position and lighting condition.}
To further validate the generalization ability, we benchmark MPI with other literature.
% under object position and lighting condition variation. 
%
We select ``put banana into basket'' as the ablation task and collect 100 demonstration trajectories, where the banana tails are randomly positioned in a 16cm × 16cm rectangular region. 
Evaluations are conducted at ten marked locations to ensure consistency and fairness across different methods. 
For instance, we regulate the tails of bananas pointing at ten marked points during inference.
% to ensure a fair comparison to prove the generalization capability.
%
Detailed settings are depicted in the Appendix. 
%
% Evaluations are conducted at 10 marked locations to ensure consistency and fairness across different methods. 
The success rates are reported in \Cref{table:generalization_position_lighting}. Our approach surpasses
% beats 
Voltron by a notable margin regardless of object position variation or lighting condition variation. Note that the prevailing alternative, MVP, 
% the leading previous method in our , 
still falls short 
% of our method 
against these variations
% variant object position and lighting conditions
in the real-robot setting.

\subsubsection{2) Franka Kitchen}
\label{sec:exp_franka}

\paragraph{Motivation.} Previous studies \citep{radosavovic2023mvp,nair2023r3m,karamcheti2023voltron} have established imitation learning for visuomotor control in simulation as the standard evaluation method. This enables direct comparisons with prior works and focuses on assessing the sample-efficient generalization of visual representations and their impact on learning policies from limited demonstrations. We conduct this evaluation to compare the capabilities of different representations in acquiring both the knowledge of ``where-to-interact'' and ``how-to-interact'' in complex simulation environments.

\begin{figure}[t!]
    \centering
    \includegraphics[width=0.95\linewidth]{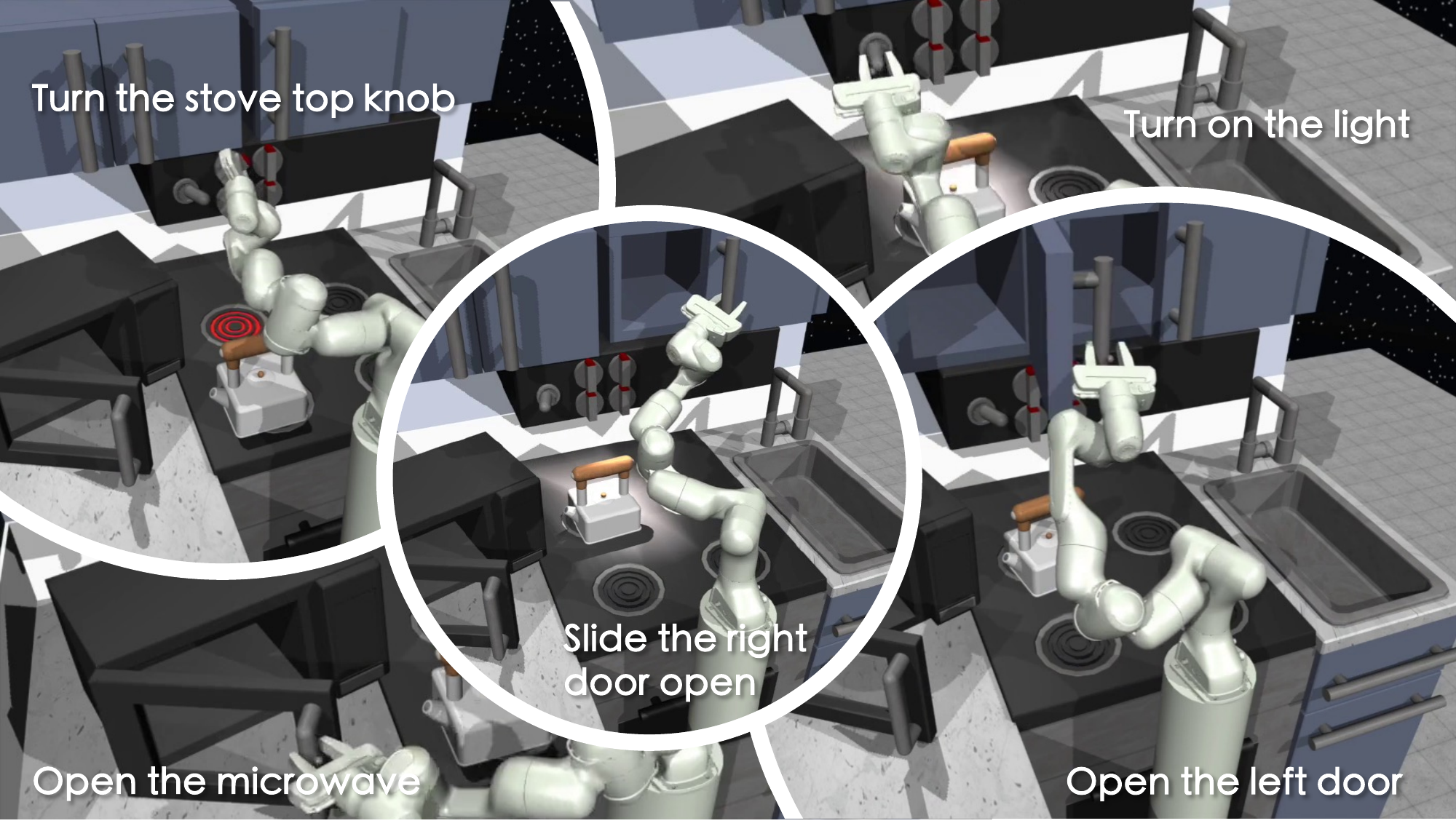}
    \caption{\textbf{Franka Kitchen simulation environment} defined by \citet{nair2023r3m}. In this environment, we include tasks of turning the stovetop knob, opening the microwave, sliding the right door open, turning on the light, and opening the left door. All tasks are trained with 25 demonstrations.}\label{fig:franka}
    % \vspace{-2pt}
\end{figure}

\paragraph{Evaluation Details.}
For the complex simulation environment, we adopt the policy learning tasks depicted in \cref{fig:franka} within the Franka Kitchen simulation environment as defined by \citet{nair2023r3m}. This environment consists of five distinct tasks, each observed from two camera viewpoints. To predict 9-DoF joint velocities (7 joints and 2 grippers) based on visual representations and proprioceptive states (\textit{i.e.}, joint velocities), we train shallow MLP policy networks for methods with a ResNet50 backbone and modified multiheaded attention pooling~(MAP)~\cite{lee2019set} for those with a ViT backbone. Following the means of \citet{nair2023r3m} and \citet{karamcheti2023voltron}, we calculate the average success rates for each setting across the 5 tasks, 2 viewpoints, and 3 random seeds. We train a separate policy head for each task to imitate RL experts and use \citet{nair2023r3m}'s codebase. Further discussions about the structure of policy networks are provided in the Appendix.

\paragraph{Experimental Results.} The results of single-task visuomotor control in the Franka kitchen are listed in \Cref{table:franka_kitchen}. Notably, representation learning frameworks tailored for robot manipulation exhibit a discernible superiority over two widely adopted visual pre-training methods in the realm of computer vision: ImageNet~\cite{deng2009imagenet} classification and CLIP~\cite{radford2021learning} contrastive pre-training. While prior methodologies demonstrate comparable performance, our implementations, built upon both smaller and larger variants of visual backbone, showcase a significant advantage over our counterparts. In particular, MPI with the ViT-Small architecture yields a noteworthy enhancement of \textbf{+6\%} in average success rate compared to the leading precedent Voltron~\cite{karamcheti2023voltron}. This improvement becomes even more pronounced with the utilization of the ViT-Base backbone, resulting in an escalation of \textbf{+7.2\%}. Furthermore, our approach consistently shows optimal or near-optimal performance across all tasks, highlighting its sustained advantage in handling the intricate scenarios encountered in the Franka Kitchen.

\paragraph{Robustness Evaluation.}
We upgrade the Franka Kitchen simulation suite to the improved setting in Burns \textit{et al.}~\cite{burns2023Robust_Representations}, which involves robustness evaluation under varying background distraction and lighting conditions. The comparisons 
% of robustness 
% across DINO, R3M, MVP, Voltron, and MPI is 
are shown in \Cref{table:franka_robustness_eval}. R3M, learning with multimodal contrastive objectives, leans towards extracting high-level semantic features, which leads to a dramatic performance loss under distractions. MPI shows superior generalization ability, where it outperforms the previous leading option MVP by 3\% on average and boosts the success rate approximately twice in challenging darker lighting conditions.

\begin{table*}[t!]
  \small
  \centering
  \caption{\textbf{Results of single-task visuomotor control on Meta-World simulation environment.} We report the success rate~(\%) over 50 randomly sampled trajectories. The best results are \textbf{bolded} and the second highest are \underline{underlined}. MPI showcases exemplary performance across three tasks, exhibiting a superior average success rate in comparison to prior methods.}
  \label{table:meta_world}
  \setlength{\tabcolsep}{2.3mm}{
    \begin{tabular}{l | cc | ccccc | c}
    \toprule
    Method &  Backbone & Param. & Assemble & Pick \& Place & Press Button & Open Drawer & Hammer & Average\\
    % \cmidrule{1-9}
    \midrule

    R3M~\cite{nair2023r3m}  & ResNet50 & 25.6M & \bf{94.0} & 60.3 & \underline{66.3} & 100 & 93.7 & 82.9 \\
    MVP~\cite{radosavovic2023mvp}    & ViT-Base & 86M & \underline{82.7} & \bf{82.0} & 62.7 & 100 & \underline{95.7} & \underline{84.6}\\
    Voltron~\cite{karamcheti2023voltron} & ViT-Small & 22M & 72.3 & 57.3 & 30.7 & 100 & 83.0 & 68.7\\

    \midrule
    \baseline{\modelname (Ours)}               & \baseline{ViT-Small}  & 
     \baseline{22M} & \baseline{69.0} & \baseline{\underline{64.0}} & \baseline{\bf{98.7}} & \baseline{\bf{100}} & \baseline{\bf{96.0}} & \baseline{\bf{85.7}}\\  
     
    % \baseline{Ours}               & \baseline{ViT-Base}  & 
    %  \baseline{86M} & \\     
    \bottomrule
    \end{tabular}}
\end{table*}

\subsubsection{3) Meta-World}
\label{sec:exp_metaworld}

\paragraph{Motivation.} 
% Mastering the art of a visuomotor control task
% goes beyond mere knowledge of “where to interact.” Under-
% standing “how to interact” allows for the integration of implicit
% action priors into subsequent policy heads.
While the scene comprises only a singular object, Meta-World~\cite{yu2020metaworld} introduces random variations in the location of the target interaction object during each validation, providing a complementary aspect to the fixed scenes presented in Franka Kitchen. Understanding ``where-to-interact'' prior to taking action enables the incorporation of implicit action priors into subsequent policy learning processes. Evaluations conducted on Meta-World emphasize the essential role of visual representations in policy learning.

\paragraph{Evaluation Details.} We adopt the tasks defined in Meta-World, which involves assembling a ring onto a peg, picking and placing a block between bins, pushing a button, opening a drawer, and hammering a nail. Following the setup introduced in \cref{sec:exp_franka}, we train shallow MLP policy heads with 25 demonstrations separately for each task using behavioral cloning. The average success rates for each setting, encompassing the 5 specified tasks, 2 viewpoints, and 3 random seeds, are calculated to obtain the results.

\paragraph{Experimental Results} for visuomotor control on Meta-World are presented in~\Cref{table:meta_world}. Remarkably, the smaller variant of MPI achieves the highest success rate, surpassing MVP~\cite{radosavovic2023mvp} with visual backbones approximately four times larger than ours. In comparison to Voltron~\cite{karamcheti2023voltron} which also uses ViT-Small, our lead extends even further to \textbf{17\%}. The superior performance in Meta-World highlights the effective generalization of the positional object movement achieved by \modelname.

\subsection{Evaluations on Referring Expression Grounding}
\label{sec:exp_REG}

\paragraph{Motivation.} The precise understanding of ``where-to-interact'' serves as a prerequisite for visuomotor control. It necessitates models to capture object-centric priors and high-level semantics. This entails considering interrelated properties such as color, light, and spatial relationships within the context of language expressions~\cite{karamcheti2023voltron}. Referring Expression Grounding~(R.E.G.) aims to predict the bounding box of an object in a cluttered scene based on a given language expression. The performance on this task is language-dependent, allowing us to assess the impact of pre-training with language instructions.

\begin{table}[t]
  \small
  \centering
  \caption{\textbf{Results of Referring Expression Grounding.} We report results of smaller variants~(ResNet50 for R3M, and ViT-Small for MVP, Voltron and ours), measured by average precision~(\%) under three IoU thresholds. $^{*}$We leverage the aggregated visual embedding $\Tilde{v}$ from the encoder. MPI yields the best detection results regardless of employing full-length visual embeddings or adopting aggregated embedding.
  % Detailed visual embedding dimensions are listed.
  }
  \label{table:refer}
  \setlength{\tabcolsep}{3mm}{
  \begin{tabular}{l| l | c c c}
    \toprule
         \multirow{2}{*}{Method} & \multirow{2}{*}{Embedding} &  \multicolumn{3}{c}{Average Precision~(AP)} \\
         % \cmidrule{3-5}
         & & @0.25 & @0.5 & @0.75 \\
         \midrule
        R3M~\cite{nair2023r3m} & $\mathbb{R}^{2048}$ & 85.27 & 71.79 & 42.66\\
        % MVP* &$ \mathbb{R}^{197 \times 384}$ & 94.43 & 88.98 &67.52 \\
        MVP~\cite{radosavovic2023mvp} & $\mathbb{R}^{384}$ & 93.07 &85.32 &60.37\\
        Voltron~\cite{karamcheti2023voltron} &$\mathbb{R}^{196 \times 384}$ & 92.93& 84.70& 57.61\\
        \midrule
        \baseline{\modelname (Ours)$^{*}$}  & \baseline{$\mathbb{R}^{384}$} &\baseline{\bf{96.29}} &\baseline{\bf{92.10}} &\baseline{71.87}\\
        \baseline{\modelname (Ours)} & \baseline{$\mathbb{R}^{196 \times 384}$} & \baseline{96.04}& \baseline{92.05} & \baseline{\bf{74.40}} \\
    \bottomrule
    \end{tabular}}
\end{table}

\paragraph{Evaluation Details.} We use the OCID-Ref Dataset~\cite{wang2021ocid}, which consists of scenes representing typical robotics environments. 
As language plays a crucial role in this task, we enhance the visual embedding by employing the DistillBERT model~\cite{sanh2019distilbert} to incorporate language information into R3M and MVP that originally lack a language encoder. 
By combining the visual embedding with the dimensions specified in~\Cref{table:refer} and the language embedding from DistillBERT, we ensure a standardized approach for a fair comparison. To extract features from the concatenation of visual and language embeddings, we introduce an MAP block~\cite{lee2019set} that is learned from the task data. Further details regarding adaptation procedures and additional analysis are provided in the Appendix.

\paragraph{Experimental Results} are in~\Cref{table:refer}. Despite its effectiveness for visuomotor control, R3M~\cite{nair2023r3m} struggles to accurately locate objects based on language descriptions, as evident from its modest 42.66\% AP under 0.75 IoU. This limitation may stem from its pre-training objective, which exclusively relies on contrastive loss to acquire high-level semantic features, with the loss of low-level localization features after global average pooling. In contrast, our method, with embeddings of the same dimension, achieves superior AP compared to the previous leading method MVP~\cite{radosavovic2023mvp}. We observe improvements of \textbf{+3.22\%}, \textbf{+6.78\%}, and \textbf{+11.5\%} under three different IoUs. In addition, the token aggregator learned from pre-training enables the model to achieve comparable or even better performance with fewer embeddings, demonstrating the effectiveness of our interaction-oriented learning target. 
% More detailed settings and results are given in the Appendix.


\begin{table*}[!t]
% \vspace{-2em}
\caption{
\textbf{
% Detailed 
Ablation studies and further comparisons.} For ablation analysis on real-world robot experiments, we select picking up the banana and placing it into the basket as our ablation tasks. We evaluate the task ten times and report the success rate (\%). Moreover, we report the average success rate (\%) on Franka Kitchen and the Average Precision (\%) @ 0.5 IoU for the Refering Expression Grounding (R.E.G.) task. We report the results on Franka Kitchen with a fixed random seed.}
\centering
% #################################################
% Ablation study of input frames
% #################################################
\hspace{-3mm}
\subfloat[
\textbf{Ablation of input frames.} The input selection is subjected to the Bernoulli distribution characterized by $p$ (in~\cref{eq:input}). Our pre-training methodology, employing the keyframes that signify states of interaction, exhibits heightened efficacy in comparison to the utilization of randomly selected frames.\label{table:ablation_pretain_paradigm}
]{
\begin{minipage}{0.47\linewidth}{
\begin{center}
\tablestyle{2.0pt}{1.0}
%\hspace{-10mm}
\begin{tabular}{x{45}|x{45}|x{44}x{43}x{43}}%{ccc|cc}
    \toprule 
    \multicolumn{2}{c|}{\multirow{2}{*}{Input Frames}} & \multirow{2}{*}{{\makecell[c]{Real-world\\Robot}}} & \multirow{2}{*}{\makecell[c]{Franka\\Kitchen}} & 
    \multirow{2}{*}{\makecell[c]{R.E.G.\\@0.5IoU}} \\
    \multicolumn{2}{c|}{} & \\
    \midrule
     \multicolumn{2}{c|}{Sequential Frames} & {40} & 74.0 & 90.57 \\
     \midrule
      \multirow{3}{*}{\makecell[c]{Key Frames\\(Ours)}}& $p=0$  & {40} &76.4 &91.73 \\
      & \baseline{$p=0.5$} & \baseline{\bf{{60}}} &   \baseline{\bf{79.2}} & \baseline{\bf{92.04}} \\
      & $p=1$ & {\bf{60}} &   78.8 & 91.77\\
    \bottomrule
  \end{tabular}
\end{center}
}
\end{minipage}
}
%#################################################
% Ablations on decoder design.
%#################################################
\hspace{3mm}
\subfloat[
\textbf{Ablation of decoder design.} \textit{Prediction} and \textit{Detection} represent the decoder blocks as introduced in~\cref{fig:pipeline}. Both decoder blocks, targeting the holistic understanding of ``how-to-interact'' and ``where-to-interact'' within the interaction process, contribute indispensably to the improved representation learning and downstream performance.\label{table:interaction}
]{
\begin{minipage}{0.47\linewidth}{
\begin{center}
\tablestyle{2.0pt}{1.0}
% \hspace{-0.5cm}
\begin{tabular}{x{45}x{45}|x{44}x{43}x{43}}
    \toprule 
    \multicolumn{2}{c|}{Decoder}& \multirow{2}{*}{{\makecell[c]{Real-world\\Robot}}} &
     \multirow{2}{*}{\makecell[c]{Franka\\Kitchen}}& \multirow{2}{*}{\makecell[c]{R.E.G.\\@0.5IoU}} \\
    % \cmidrule[0.5pt]{1-2} 
    Prediction & Detection &   &\\
    \midrule
    &  &{30}  & 62.4  & 63.94 \\
    $\checkmark$&  &{50}  &  74.8 & 90.58\\
    &$\checkmark$&{50}  &  75.2 & 90.72 \\
    \midrule
    \baseline{$\checkmark$}&\baseline{$\checkmark$} & \baseline{\bf{{60}}} &\baseline{\bf{79.2}} & \baseline{\bf{92.04}} \\
     % \midrule
     
    \bottomrule
\end{tabular}
% \hspace{-10mm}
\end{center}
}\end{minipage}
}
\vspace{5pt}
% #################################################
% Ablation study of causality modeling mechanism.
% #################################################
\hspace{-3mm}
\subfloat[\textbf{Ablation of causality modeling mechanism.} \textit{Dual-frame} indicates that the visual embedding of the two frames is fed to the decoder in parallel during pre-training. \label{table:ablation_temporal}
%\hspace{-10mm}
]{
\begin{minipage}{0.47\linewidth}{
\begin{center}
\tablestyle{2.0pt}{1.0}
% \hspace{-10mm}
\begin{tabular}{x{90}|x{44}x{43}x{43}}%{l | c c }
    \toprule
    \multirow{2}{*}{Causality Modeling} &  \multirow{2}{*}{\makecell[c]{Franka\\Kitchen}} & \multirow{2}{*}{{\makecell[c]{Robustness\\Evaluation}}} & \multirow{2}{*}{\makecell[c]{R.E.G.\\@0.5IoU}}   \\
    & & & \\
    \midrule
    %\midrule
    w/o (Dual-frame) &77.6 &{47.1}  & 91.93\\
    % Temporal-wise Mean & 90.72 & 78.8\\
    \baseline{Deformable Attn.}& \baseline{\bf{79.2}} & \baseline{\bf{{49.7}}} &\baseline{\bf{92.04}} \\
    \bottomrule
  \end{tabular}
\end{center}
}
\end{minipage}
}
% #################################################
% Ablation study on input modality.
% #################################################
\hspace{3mm}
\subfloat[
\textbf{Ablation of input modality.} For \textit{Vision-only}, we discard language prior in policy learning.
Language modality is beneficial while competitive performance is still achieved without language. \label{table:ablation_language}
]{
\begin{minipage}{0.47\linewidth}{
\begin{center}
\tablestyle{2.0pt}{1.0}
% \hspace{-10mm}
\begin{tabular}{x{90}|x{65}x{65}}
\toprule 
 \multirow{2}{*}{Input Modality} & \multirow{2}{*}{\makecell[c]{Franka\\Kitchen}} & \multirow{2}{*}{{\makecell[c]{Robustness\\Evaluation}}}\\
 & & \\
\midrule
Vision-only & 77.2 & {44.3}\\
\baseline{Vision + Language} & \baseline{\textbf{79.2}} & \baseline{\textbf{{49.7}}}\\
\bottomrule
\end{tabular}
\end{center}
}
\end{minipage}
}
\label{tab:ablation}
\underfigtab
\end{table*}

\subsection{Ablations \& Further Analysis}
\label{sec:exp_ablation}

% All experiments in this section, unless specified otherwise, are conducted with the ViT-Small variant. Notably, in contrast to the results shown in~\Cref{table:franka_kitchen}, we report the average success rate on Franka Kitchen with a fixed random seed.

\paragraph{Input Frames.} 
\label{ablation:pretrain_paradigm}
\modelname uses three keyframes from a video clip, distinguishing it from conventional pre-training approaches based on video prediction.
To validate this design, we evaluate representation learning with three consecutive frames, as presented in the top row of~\Cref{table:ablation_pretain_paradigm}. Specifically, the selection of data frames centers around the transition frame $F_{\text{trans}}$ or the final frame  $F_{\text{final}}$ with corresponding box annotations. This is to support the learning of interaction object detection and maintain consistency with our designated learning objectives. 

% To alleviate potential performance variations stemming from the temporal gap between frames, we further extend the time span to $(T_{\text{final}} - T_{\text{init}}) / 2$, where $T_{\text{final}}$ and $T_{\text{init}}$ denote the timestamps corresponding to the final and initial frames.

As outlined in~\Cref{table:ablation_pretain_paradigm}, the performance with sequential frames is inferior to our pre-training paradigm, which is grounded upon key interaction states. Furthermore, the alternating use of transition and final frames as prediction targets~(with $p$ set to 0.5 in~\cref{eq:input}) enhances the model's comprehensive understanding of the interaction process, leading to further advancements in performance.

\paragraph{Decoder Design.}
% {\color{red}How-to-interact \& Where-to-interact}
As depicted in the right section of \cref{fig:pipeline}, we employ a pair of transformer decoder blocks designated as the prediction transformer and detection transformer. Each decoder is assigned a specific role: the Prediction Transformer predicts the unseen interaction state, while the Detection Transformer infers the interaction object in the unseen frame. In~\Cref{table:interaction}, we investigate the functions performed by each decoder component. In contrast to employing transformer-based decoders, the first row of~\Cref{table:interaction} showcases an alternative approach where we utilize a 3-layer MLP to directly decode the target frame from encoder embeddings. The utilization of either of the introduced decoders leads to noteworthy improvements in both tasks, evidenced by an increase of over +12\% average success rate on Franka Kitchen and a +26\% AP improvement in the R.E.G. task. The synergistic integration of the prediction and detection transformer contributes to enhanced representation learning, ultimately achieving optimal performance and demonstrating the necessity of these two interaction-oriented designs in our framework.

\paragraph{Causality Modeling.}
For the multi-modal encoder, we design a causality modeling mechanism built upon the deformable attention~\cite{zhu2021deformable} to capture the causality relationships between two input states during pre-training. As shown in \Cref{table:ablation_temporal}, excluding this module results in a decline in performance. The comparable performance in R.E.G. can be attributed to the task's focus on static scene detection, which does not strictly require an understanding of interaction dynamics. It is important to note that 
% our primary focus is on the contribution of this mechanism to representation learning during pre-training, while 
the image features from the ViT are simply replicated to constitute inputs for this module in all downstream tasks. The additional parameters~(\textit{e.g.}, 0.3M for ViT-Small) and computational load introduced are negligible.

\paragraph{Involvement of Language.} Our pre-training pipeline is language-conditioned. Nevertheless, we have decoupled the visual and language branches in our encoder. Therefore, the language modality is not requisite in downstream tasks. As shown in \Cref{table:ablation_language}, the inclusion of language leads to a +2\% increase in success rate in Franka Kitchen. This improvement primarily originates from the ability of textual features to highlight the most relevant visual features for the task. 
% Furthermore, it should be 
Note that notwithstanding the absence of language, \modelname still achieves higher performance compared to other representation learning frameworks evaluated in \Cref{table:franka_kitchen}.

\begin{figure}[t!]
    \centering
    \includegraphics[width=1.0\linewidth]{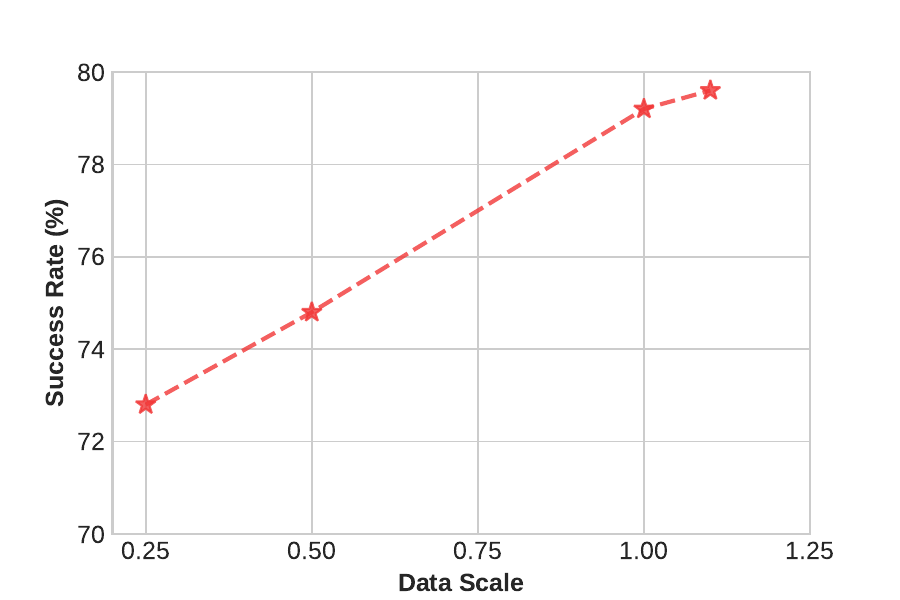}
    \caption{\textbf{Influence of pre-training data scale.} We test on the single-task visuomotor control task. \modelname demonstrates improvement with scaled pre-training data, holding the potential for further performance enhancement with large-scale data.
    }
    \label{fig:pretrain_data_scale}
\end{figure}

\paragraph{Data Regime.} We conduct ablation experiments on data scale, where we pre-train \modelname using data volumes of 25\%, 50\%, 100\%, and 110\%, respectively. Subsequently, we evaluate them in the Franka Kitchen environment. Results in \cref{fig:pretrain_data_scale} demonstrate that increasing the scale of pre-training data correlates with improved performance in downstream tasks.

\section{Conclusions} 
\label{sec:conclusion}

In this work, we present \modelname, an interaction-oriented representation learning framework to empower manipulation tasks.  
By predicting transition states based on initial and final states, \modelname enhances the model's understanding of interactive dynamics.
Through extensive experiments, we demonstrate that our method achieves state-of-the-art performance across a diverse range of downstream tasks.

\paragraph{Limitations and Future Work.} 
Our framework by far utilizes explicit annotations~(\textit{i.e.,} keyframes, languages, and bounding boxes for interaction object) provided in the Ego4D-HoI~\cite{grauman2022ego4d} dataset. This could limit the applicability of our methods to broader datasets. Nonetheless, it is feasible to leverage some well-established vision tools~\cite{liu2023llava, chen2023affordance} to collect annotations cheaply and scale up the pre-training dataset with web-crawled images.
Additionally, the policy model employed in this study, which relies on behavior cloning from a limited amount of expert demonstrations, restricts the generalization of the model to different scenarios.
Besides, the lack of a large language model constrains the model's ability to perform long-horizon planning and causal reasoning.
Exploring the incorporation of the pre-trained model from \modelname into vision-language models to tackle long-horizon interaction-related tasks represents an interesting direction for future research.

\section*{Acknowledgments}
This work was supported by National Key R\&D Program of China (2022ZD0160104), NSFC (62206172), Shanghai Committee of Science and Technology (23YF1462000), and China Postdoctoral Science Foundation (2023M741848). 

% \clearpage
% \newpage
\bibliographystyle{plainnat}
\bibliography{bibliography_short, bibliography_custom}

\clearpage
\newpage

%\appendix
%\input{paper-template-latex/rebuttal_response}
% \maketitlesupplementary
\title{
Learning Manipulation by Predicting Interaction \\
\textit{Supplementary Material}
}

% % You will get a Paper-ID when submitting a pdf file to the conference system
% \author{Author Names Omitted for Anonymous Review. Paper-ID 
% [6]}

\renewcommand{\thetable}{A-\Roman{table}}
\renewcommand{\thefigure}{A-\arabic{figure}}
\setcounter{figure}{0}
\setcounter{table}{0}  

\onecolumn
\author{}
\maketitle

\appendix

\medskip
\noindent 
Within the appendices presented below, we first conduct in-depth discussions about a range of motivating questions.
Secondly, we provide additional ablation studies to support our viewpoints in our responses to the motivating questions. 
Subsequently, we offer extra details regarding the pre-training process.
Following this, 
%adhering to the sequential arrangement in Sec.~\textcolor{red}{\uppercase\expandafter{\romannumeral 4}} in the main paper, 
we furnish supplementary elaborations pertaining to the real-world robot experiments, Franka kitchen experiments, and referring expression grounding. 
%
%Following this, we proceed to visually represent the feature representation and conduct a thorough saliency analysis. 
%
Lastly, we introduce additional relevant works involving vision-language models in robotics.
Please note that we enclose a demonstration video in the attached files, to present the efficacy and robustness of our approach.

Summaries for specific topics are presented below:

\medskip
\noindent \rule{\linewidth}{0.5pt}
\medskip

\noindent \hyperref[sec:motivating_questions]{{\color{LightSlateBlue}Appendix A -- \textit{Motivating Questions}}}

\begin{quote}
    \medskip
    We compile a list of ``motivating'' questions potentially raised from reading the main text, and we provide detailed explanations for each query. For certain questions, our answers link to the further experiments in the appendices.
\end{quote}

\bigskip

\noindent \hyperref[sec:additional_ablationstudies]{{\color{LightSlateBlue}Appendix B -- \textit{Additional Ablation Studies}}}

\begin{quote}
    \medskip
    We provide additional ablation studies on Detection Transformer as well as the choice of pre-training and downstream adaptation paradigms. Specifically, the first study demonstrates the advantage of inferring the interaction object positions in unseen frames, and the second study validates the soundness of encoder pre-training and freezing. Further discussions of pre-training data regime are also provided. 
\end{quote}

\bigskip

\noindent \hyperref[sec:Extra_Pretraining_Details]{{\color{LightSlateBlue}Appendix C -- \textit{Supplementary Pre-training Details}}}

\begin{quote}
    \medskip
    We present the pre-training specifics concerning the pre-training paradigm and image processing.
\end{quote}

\bigskip

\noindent \hyperref[sec:details_real_robots]{{\color{LightSlateBlue}Appendix D -- \textit{Experimental Details of Real-world Robots}}}

\begin{quote}
    \medskip
    We provide extra experiment configuration details and use a Sankey diagram to display fine-grained failure causes. Based on the analysis of the Sankey diagram, it can be deduced that the primary causes of failures are attributed to occlusion, incorrect localization, and improper grasping.
\end{quote}

\bigskip

\noindent \hyperref[sec:details_franka_kitchen]{{\color{LightSlateBlue}Appendix E -- \textit{In-Depth Analysis into Franka Kitchen}}}

\begin{quote}
    \medskip
    We illustrate our proposed feature aggregation strategy for visuomotor control tasks, and analyze the impact of this strategy on the Franka kitchen experiment.
    The experimental results showcase the superior performance of our proposed aggregation approach in comparison to preceding aggregation strategies. 

\end{quote}

\bigskip

\noindent \hyperref[sec:details_reg]{{\color{LightSlateBlue}Appendix F -- \textit{{A Closer Look to Referring Expression Grounding}}}}

\begin{quote}
    \medskip
    We provide detailed evaluation information for the Referring Expression Grounding task, and offer specific replication guidelines for each method covered.
\end{quote}

\bigskip

% \noindent \hyperref[sec:visualization]{Appendix E -- \textit{Visualization}}

% \begin{quote}
%     \medskip
%     By visualizing features for saliency analysis, we showcase the model's attention towards interaction objects in complex scenes.
% \end{quote}

% \bigskip

\noindent \hyperref[sec:more_related_works]{{\color{LightSlateBlue}Appendix G -- \textit{More Related Works}}}

\begin{quote}
    \medskip
    We introduce a series of studies highlighting the utilization of visual language models in robotics, along with a discussion of our connection to previous research in this field.
\end{quote}

\bigskip

\clearpage
\newpage

% \section*{Overview}
% \startcontents
% {
% \hypersetup{linkcolor=black}
% \printcontents{}{1}{\noindent\textbf{\Large Appendix}\vskip5pt
% }

% \maketitlesupplementary
% \maketitle

\newpage

%\section{Appendix}
\subsection{Motivating Questions}
\label{sec:motivating_questions}

To facilitate a clearer understanding of our work, we supplement with intuitive questions as follows.

\bigskip

\textbf{Q1:} \textit{Under the condition of pre-training with a smaller amount of data, \modelname exhibits superior performance in downstream tasks compared to R3M, MVP, and Voltron. What is the underlying source of this improvement?}
\smallskip

Our algorithm consistently outperforms R3M~\cite{nair2023r3m}, MVP~\cite{radosavovic2023mvp}, and Voltron~\cite{karamcheti2023voltron} in various downstream tasks as presented in Table~\textcolor{red}{\uppercase\expandafter{\romannumeral 3}}-\textcolor{red}{\uppercase\expandafter{\romannumeral 6}} of the main paper.

The improvements may arise from three distinct facets: 

\begin{enumerate}
    \item MVP's approach of modeling masked images on a per-frame basis falls short in capturing essential temporal information. R3M incorporates video-language alignment. However, aligning the textual semantic feature could be achieved by simply summarizing the cues from multiple frames, without necessitating the temporal evolution modeling between frames. In contrast, our proposed interaction-oriented prediction framework captures not only the temporal evolution but also the interactive dynamics of the manipulation scenes.
    \item The Detection Transformer proposed in our framework urges the model to infer the location of the interaction object in the unseen frame. Such an object-centric design aids in accurately recognizing the target in complex scenarios.  
    \item \modelname simultaneously performs image prediction and object position inference on unseen frames. These two tasks are mutually supported, making it potentially superior in terms of perception-prediction joint reasoning.
\end{enumerate}
% 1) MVP's approach of modeling masked images on a per-frame basis falls short in capturing essential temporal information. R3M incorporates video-language alignment. Sometimes aligning the textual semantic feature can be achieved by simply summarizing the cues from multiple frames, without necessitating the temporal evolution modeling between frames. Our proposed interaction-oriented prediction framework captures not only the  temporal evolution but also the interactive dynamics of the manipulation scenes.

% 2) The Detection Transformer in our framework urges the model to infer the location of the interaction object in the unseen frame. Such object-centric design aids in accurately recognizing target in complex scenarios.  

% 3) \modelname simultaneously performs image prediction and object position inference on unseen frames. These two tasks are mutually supported, making it potentially superior to R3M and MVP in terms of perception-prediction joint reasoning.

\bigskip

\textbf{Q2:} \textit{Why is the inference of the interaction object position in the unseen frame preferred over that in the seen frame?}
\smallskip
% visible/seen, invisible/unseen

%The better performance achieved by inferring object positions in unseen frames compared to seen frames may arise from the following aspects:
The superiority of inferring object locations in unseen frames, as opposed to inferring object locations in seen frames, lies in the following aspects:
\begin{enumerate}
    \item Performing inference of interaction object positions in unseen frames not only examines the model's proficiency in fine-grained recognition and localization, but also necessitates the model's capability to deduce the temporal evolution of object positions. This demanding task considerably enhances the reasoning ability of the pre-trained encoder.
    \item Inferring object positions in unseen frames can establish more effective mutual support with the task of image prediction for unseen frames, given that the image prediction process is operated in unseen frames. 
\end{enumerate}
% 1) Performing inference of interaction object positions in unseen frames not only examines the model's proficiency in fine-grained recognition and  localization, but also necessitates the model's capacity to deduce the temporal evolution of object positions.
% %
% This demanding task considerably enhances the capabilities of the encoder.
% %
% 2) From another perspective, inferring object positions in unseen frames can establish a more effective mutual support with the task of image prediction for unseen frames.
We complete comparative experiments in \Cref{table:ablation_target_detection} to prove the effectiveness of inferring object locations in unseen frames.

% \begin{table}[h]
%   \small % \footnotesize \small \normalsize
%   \centering
%   \caption{The ablation experiments on the target frame for interaction object detection.}
%   \label{table:ablation_frame_iod}
%   \setlength{\tabcolsep}{2mm}{
%   \begin{tabular}{l|cc}
%     \toprule
%     Target frame & Franka Kitchen  & R.E.G.@0.5 IoU \\
%     \midrule
%     seen & 70.8  & 91.60 \\
%     %\midrule
%     unseen & \baseline{\bf{79.2}}& \baseline{\bf{92.04}}\\
%     \bottomrule
%   \end{tabular}}
% \end{table}

\bigskip

\textbf{Q3:} \textit{Why is the evaluation limited to a frozen encoder? Could we also compare the performance between a pre-trained frozen encoder and an encoder trained from scratch? Additionally, why is the option of fully fine-tuning the encoder not considered?}
\smallskip

Previous methods, such as R3M, MVP, and Voltron, only evaluate frozen visual representations. We adopt a consistent setting with them for a fair comparison.
In contrast to training from scratch, pre-training offers the model a favorable initialization and enables the acquisition of more generalizable representations.
Fully fine-tuning the pre-trained visual encoders with a limited amount of adaptation data may lead to severe overfitting.
Therefore, we freeze the encoder on downstream tasks and refrain from fine-tuning its weights.
We conduct additional ablative experiments in \Cref{table:ablation_finetune} to substantiate our viewpoint.

% \begin{table}[h]
%   \small % \footnotesize \small \normalsize
%   \centering
%   \caption{The ablation experiments on the effect of pre-trained and frozen encoder.}
%   \label{table:ablation_finetune}
%   \setlength{\tabcolsep}{2mm}{
%   \begin{tabular}{c|c|ccc}
%     \toprule
%     \multicolumn{2}{c}{Approach} & Pre-train & Freeze & Success Rate \\
%     \midrule
%     % 3-layer CNN &  &  & \\
%     % \midrule
%     \multicolumn{2}{c|}{train from scratch} & $\times$ & $\times$ & 71.6\\
%     \multicolumn{2}{c|}{finetune pre-trained encoder} & $\checkmark$ & $\times$ & 77.6\\
%     \multicolumn{2}{c|}{freeze pre-trained encoder}& \baseline{$\checkmark$} & \baseline{$\checkmark$} & \baseline{\textbf{79.2}}\\
%     \bottomrule
%   \end{tabular}}
% \end{table}

\bigskip

\textbf{Q4:} \textit{Why not directly adopt explicit representation for downstream tasks? How about utilizing affordances to guide training ``where-to-interact''?}
\smallskip

Instead of incorporating object position information into downstream tasks directly, we utilize the embedding that contains semantic information about object positions. Such a manner is more flexible for downstream task learning.
Affordance can also guide training ``where-to-interact'', and our framework is compatible with the design. However, to attain affordance annotations, human labeling is laborious, and current auto-labeling techniques pose challenges in ensuring accuracy, which results in the scarce open-sourced annotations on ego-centric human video datasets. In contrast, ours requirements for box annotation in key frames is less restrictive.

\clearpage
\newpage

\bigskip

\subsection{Additional Ablation Studies and Discussions}
\label{sec:additional_ablationstudies}

\paragraph{Target frame for Detection Transformer.} We perform ablation experiments to explore the optimal choice of target bounding boxes between those from the seen frames and those from the unseen frames.
As presented in~\Cref{table:ablation_target_detection}, detecting object locations in seen frames, compared to inferring in unseen frames, results in a significant performance decline of \textbf{-4.8\%} in the Franka Kitchen task. As discussed in \cref{sec:motivating_questions}~(\textcolor{red}{Q2}), the approach of inferring in unseen frames enhances the model’s capability to deduce the temporal evolution of object positions, and it can establish a more effective symbiotic relationship with the Prediction Transformer. These two aspects contribute to its superiority over detecting in seen frames.

%\paragraph{Encoder pre-training and freezing.}
\paragraph{Downstream adaptation paradigm.}
By default, we pre-train the model on a large-scale ego-centric video dataset and subsequently freeze the encoder for downstream tasks. To showcase the advantages of pre-training, we assess the model's performance when fully training from scratch. The results shown in~\Cref{table:ablation_finetune} reveal a substantial performance degradation by \textbf{-7.6\%} for models without pre-training. In addition, we explore further fine-tuning the pre-trained encoder on downstream data, which leads to a performance decrease of \textbf{-1.6\%}. 
These findings validate the soundness of our experimental setup. The visual encoder benefits from 
pre-training, and should be frozen in downstream tasks to mitigate overfitting.

\begin{table*}[t!]
% \vspace{-2em}
\caption{
\textbf{Additional ablation studies.} The experiments are conducted on Franka Kitchen environment, and the experimental configuration is the same as \Cref{tab:ablation} in the main paper. Default settings are marked in \colorbox{baselinecolor}{gray}.}
\centering
% #################################################
% Ablation study of causality modeling mechanism.
% #################################################
\hspace{-3mm}
\subfloat[The ablation experiments on the target frame for interaction object detection. \label{table:ablation_target_detection}
%\hspace{-10mm}
]{
\begin{minipage}{0.46\linewidth}{
\begin{center}
\tablestyle{2.0pt}{1.0}
% \hspace{-10mm}
\begin{tabular}{x{90}|x{130}}%{l | c c }
    \toprule
    \multirow{2}{*}{Target frame} & \multirow{2}{*}{Success Rate} \\
     & \\
    \midrule
    Seen & 74.4 \\
    %\midrule
    \baseline{Unseen} & \baseline{\bf{79.2}}\\
    \bottomrule
  \end{tabular}
\end{center}
}
\end{minipage}
}
% #################################################
% choice of pre-training and downstream adaptation paradigm.
% #################################################
\hspace{1mm}
\subfloat[
The ablation experiments on the choice of pre-training and downstream adaptation paradigm. \label{table:ablation_finetune}
]{
\begin{minipage}{0.46\linewidth}{
\begin{center}
\tablestyle{2.0pt}{1.0}
% \hspace{-10mm}
\begin{tabular}{x{35}|x{35}|x{40}x{40}|x{50}}
    \toprule
    \multicolumn{2}{c|}{Approach} & Pre-train & Freeze & Success Rate \\
    \midrule
    % 3-layer CNN &  &  & \\
    % \midrule
    \multicolumn{2}{c|}{train from scratch} & $\times$ & $\times$ & 71.6\\
    \multicolumn{2}{c|}{finetune pre-trained encoder} & $\checkmark$ & $\times$ & 77.6\\
    \multicolumn{2}{c|}{\baseline{freeze pre-trained encoder}}& \baseline{$\checkmark$} & \baseline{$\checkmark$} & \baseline{\textbf{79.2}}\\
    \bottomrule
  \end{tabular}
\end{center}
}
\end{minipage}
}
\label{tab:additional_ablation}
\underfigtab
\vspace{10pt}
\end{table*}

\paragraph{Pre-training data regime.} The data regimes leveraged by the pre-training methods R3M, MVP, Voltron, and our method \modelname are different. In \Cref{table:data_regime} below, we summarize the adopted pre-training datasets and their corresponding data scale.
It can be observed that the data regime of the dataset employed by our method is \textbf{smaller} in scale compared to the other three methods, yet our method yields superior performance outcomes.

\begin{table*}[h!]
  \small
  \centering
  \caption{\textbf{Summary of data regimes leveraged by various pre-training approaches.}
  }
  \label{table:data_regime}
    \centering
    \begin{tabular}{l | c | c}
    \toprule
    Method &  Data Source & Data Scale \\
    \midrule
    R3M    & Ego4D & 3M \\
    MVP   & Ego4D, Something-Something, Epic-Kitchens, etc. &  4.5M \\
    Voltron & Something-Something & 168K \\
    \modelname (Ours)   & Ego4D Hand-Object Interaction Subset & 93K \\ 
    \bottomrule
    \end{tabular}
\end{table*}

We select the Hand-Object Interaction subset from Ego4D as our pre-training data foundation, as this dataset depicts scenarios of human hands interacting with environmental objects from an ego-centric view. We filter for data that includes deterministic actions accompanied by textual descriptions of these actions. Furthermore, samples that contain complete key frame annotations and bounding boxes of manipulated objects are selected for these key frames.

We have reproduced R3M, MVP, and Voltron on the dataset we employed, aligning all methods to the \textbf{same} pre-training data regime. The results under Franka Kitchen simulation environments are listed in \Cref{table:rebuttal_align_data_regime} below. %\cl{Their performance compared to pre-training on their own data.} 
%In comparison to the original papers' use of pre-training data, Voltron exhibits a performance improvement of approximately 3.9\% in Franka, while R3M demonstrates a noticeable decline. 
It can be observed that MPI consistently maintains superior overall performance compared to existing approaches.

 \begin{table*}[t!]
  \small
  \centering
  \caption{\textbf{Results of single-task visuomotor control on Franka Kitchen and Meta-world.}
    The success rate (\%) over 50 randomly sampled trajectories is reported. We \textbf{bold} the best result and \underline{underline} the second. With the same pre-training data regime and model size, MPI still exhibits better overall performance across two simulation environments.
  }
  \label{table:rebuttal_align_data_regime}
    \centering
    \setlength{\tabcolsep}{3mm}{
    \begin{tabular}{l|l | cc | ccccc | c}
    \toprule
    Env. & Method &  Backbone & Param. & Task 1 & Task 2 & Task 3 & Task 4 & Task 5 & Average\\
    \midrule
    \multirowcell{4}[0pt][l]{Franka\\Kitchen}& R3M   & ViT-Small & 22M & 55 & 34 & 79 & \underline{45} & 97 & 62.0\\
    &MVP   & ViT-Small & 22M & 84 & 48 & \textbf{96} & 36 & \underline{100} & 72.8\\
    &Voltron & ViT-Small & 22M & \textbf{92} & \textbf{56} & 82 & 44 & 98 & \underline{74.4}\\
    &\baseline{\modelname (Ours)}                & \baseline{ViT-Small} & \baseline{22M} &  \baseline{\underline{88}} & \baseline{\underline{54}} & \baseline{\underline{90}} & \baseline{\textbf{64}} & \baseline{\textbf{100}} & \baseline{\textbf{79.2}} \\ 
    \midrule
    \midrule
    \multirowcell{4}[0pt][l]{Meta\\World}&R3M   & ViT-Small & 22M     & \textbf{76}    &51	&\underline{89}	&100	&\textbf{99} &\underline{83.0}\\
    &MVP   & ViT-Small & 22M     & 70	&56	&86	&100	&68	&76.0\\
    &Voltron & ViT-Small & 22M   & \underline{72}	&\textbf{74}	&80	&100	&82	&81.6\\
    &\baseline{\modelname (Ours)}                & \baseline{ViT-Small} & \baseline{22M} &  \baseline{70} & \baseline{\underline{64}} & \baseline{\textbf{100}} & \baseline{\textbf{100}} & \baseline{\underline{96}} & \baseline{\textbf{86.0}} \\ 
    \bottomrule
    \end{tabular}}
\end{table*}

\begin{table}[t]
    \centering
    \small
    \caption{\textbf{Performance comparison using different annotation sources.}}
    \label{tab:auto_label}
    \setlength{\tabcolsep}{5mm}{
    \begin{tabular}{c|c|c}
    \toprule
         Annotation & Human-annotated & Auto-labeling\\
         \midrule
         Franka Kitchen Success Rate & 72.8 & 72.0\\
        \bottomrule
    \end{tabular}}
\end{table}

\paragraph{Pre-training with auto-labeled data.}
We sample 25\% of the adopted human video data and replace the human-annotated language descriptions in our pre-training dataset with auto-labeled ones generated by Vision-Language Models (LLaMA-Adapter~\cite{gao2023llamaadapterv2}). We then evaluate the model accordingly, and the results are presented in~\Cref{tab:auto_label}. Utilizing auto-labeled language descriptions led to a slight decrease in performance for our framework. 
It reveals the effectiveness of our method under automatically annotated samples.
%
%This indicates the potential for cost-effective data collection and the opportunity to enhance the performance of MPI through the use of large-scale automatically annotated datasets.

\subsection{Supplementary Pre-training Details}
\label{sec:Extra_Pretraining_Details}

\begin{figure}[t!]
    \centering
    \includegraphics[width=\linewidth]{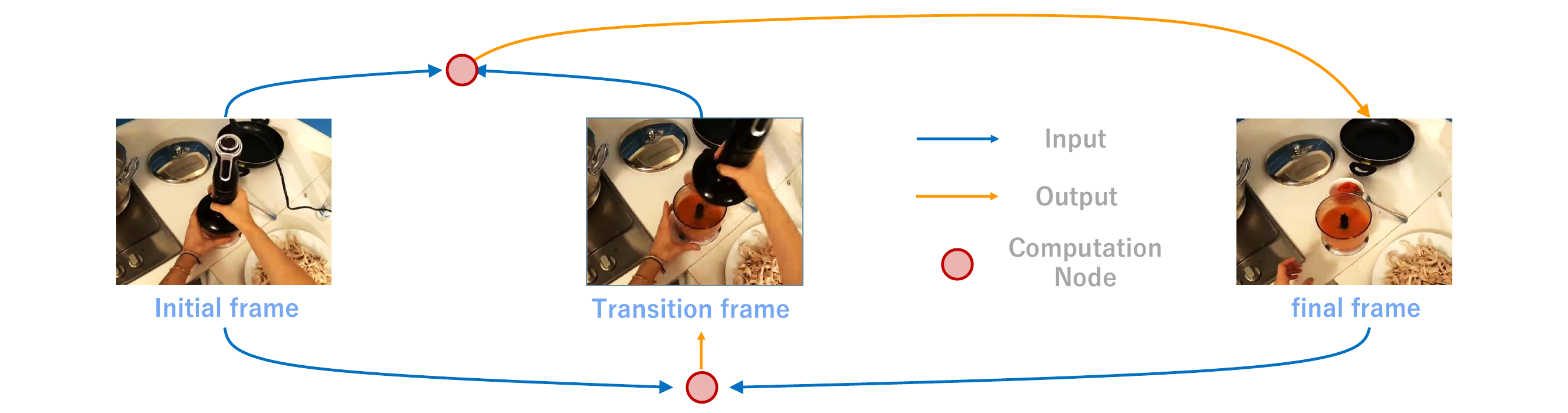}
    \caption{\textbf{Pre-training paradigm using three key frames.}}
    \label{fig:data_pipeline}
\end{figure}

The pre-training paradigm employing three keyframes is depicted in \cref{fig:data_pipeline}. As expounded in the main paper, the parameter $\alpha$ in \cref{eq:input} determines the target frame selection, whether it is the transition frame or the final frame. 
Specifically, 
when $\alpha$ is set to 1, the model predicts the transition frame and infers the interacting object position within the transition frame, given the initial and final frames. 
Conversely, when $\alpha$ is set to 0, the model predicts the final frame and infers the position of the interacting object within the final frame, given the initial and transition frames.
% for $\alpha=1$, the model predicts the transition frame and infers object positions, given the initial and final frames; for $\alpha=0$, the model predicts the final frame and infers transitioning object positions, given the initial and transition frames. 

Due to the diverse devices used for video capture in Ego4D dataset, the captured images have varying resolutions. We perform the following processing steps to standardize the input images format.
Three key frames are extracted from each video. For each frame, the longer side is cropped to match the shorter side, followed by resizing to $224 \times 224$. Finally, normalization is performed using the mean and variance of the ImageNet dataset.
%
%We adopt cosine decay with warmup as the learning rate schedule, and the warmup epoches is set to be 10.

\begin{figure}
    \centering
    \includegraphics[width=0.9\linewidth]{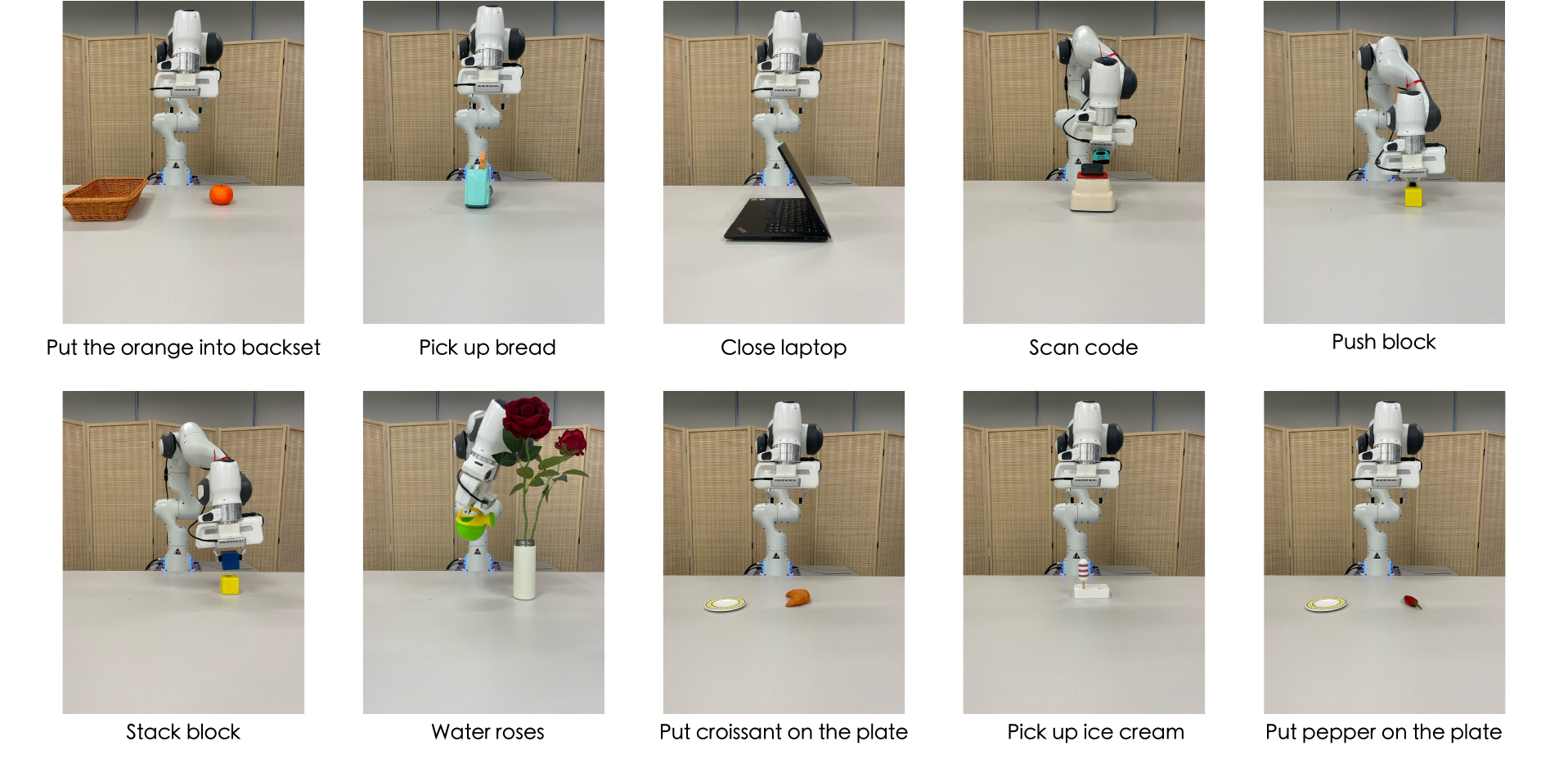}
    \caption{\textbf{A demonstration of real-world experiment setting of ten tasks conducted in clean background.}}
    \label{fig:setting_clean}
\end{figure}

\begin{figure}[tb!]
    \centering
    \includegraphics[width=0.75\linewidth]{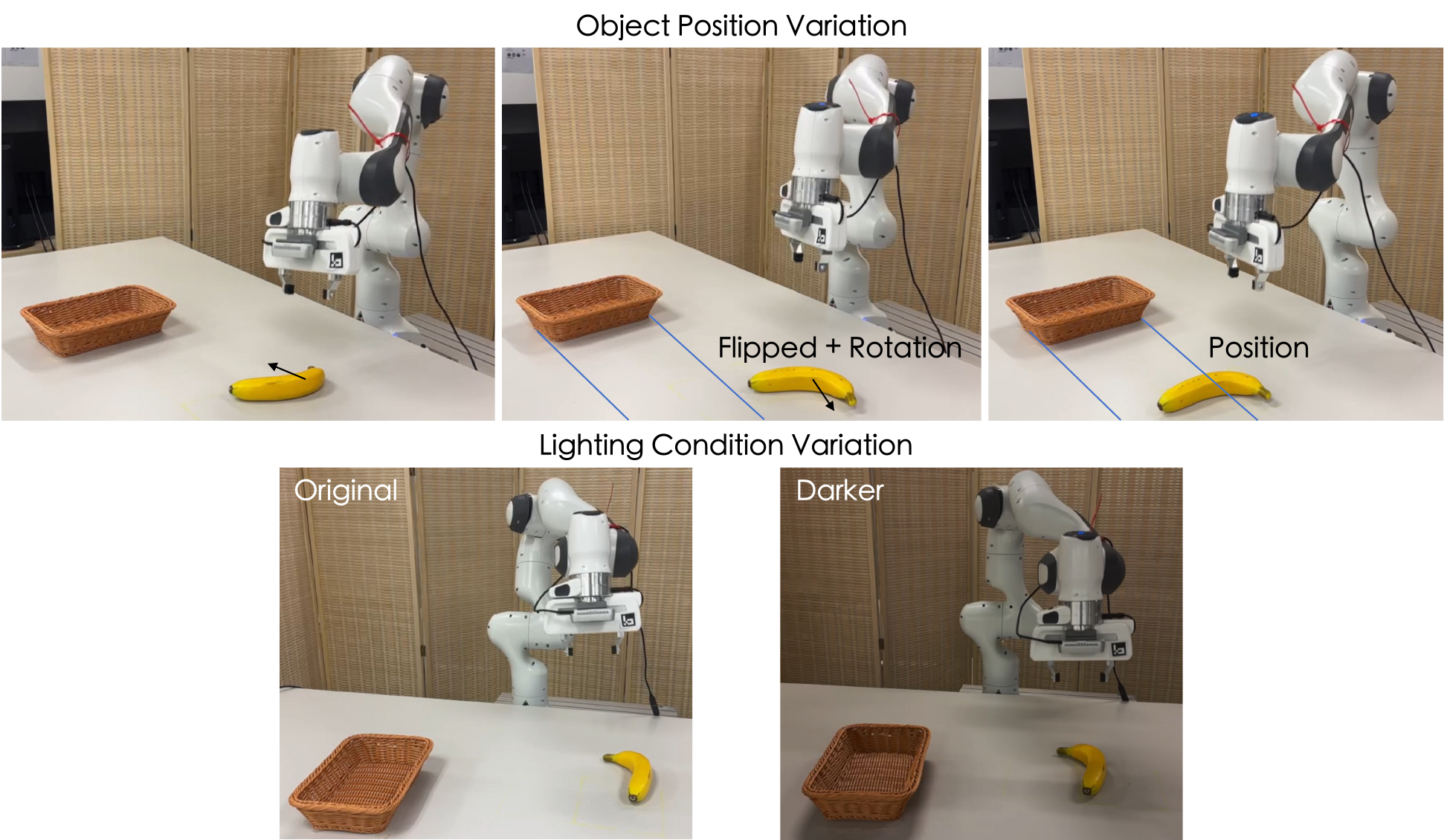} 
    \caption{\textbf{Real-world setting for the evaluation of object position and lighting condition generalization.} (a) Object position variation: the banana is randomly placed within a rectangular area during the demonstration collection stage, and at the testing stage, we marked ten spots for evaluation to ensure testing consistency between methods. (b) Lighting condition variation: the lighting in the scene becomes darker, and more shadows are present.}
    \label{fig:generaliation_setting}
\end{figure}

\subsection{Experimental Details of Real-World Robot}
\label{sec:details_real_robots}

We evaluate the applicability of our algorithm on real-world robotic systems through assessment in two distinct scenarios: five tasks within a complex kitchen environment, and ten tasks in a clean background. 
In both environments, a third-person perspective camera configuration is utilized. Specifically, the camera is installed on the rear right side in the kitchen environment and in front in the clean background environment.
The illustrations of ten manipulation tasks in the clean background are shown in~\cref{fig:setting_clean}. In contrast to the kitchen environment illustrated in \cref{fig:real-world}(a), these tasks do not involve situating the target objects within complex scenes, leading to a substantial reduction in visual interference. Nevertheless, these tasks necessitate fine-grained object recognition, alongside specialized operations such as manipulating articulated objects. Within the paradigm of imitation learning with limited demonstrations, attaining high success rates still remains a considerable challenge.

% \begin{figure}[tb!]
%     \centering
%     \includegraphics[width=0.6\linewidth]{figures/generalization_exp_setting.pdf} 
%     \caption{\revision{\textbf{Real-world setting for the evaluation of object position and lighting condition generalization.} (a) Object position variation: the banana is randomly placed within a rectangular area during the demonstration collection stage, and at the testing stage, we marked ten spots for evaluation to ensure testing consistency between methods. (b) Lighting condition variation: the lighting in the scene becomes darker, and more shadows are present.}}
%     \label{fig:generaliation_setting}
% \end{figure}

\begin{figure}
    \centering
    \includegraphics[width=.9\linewidth]{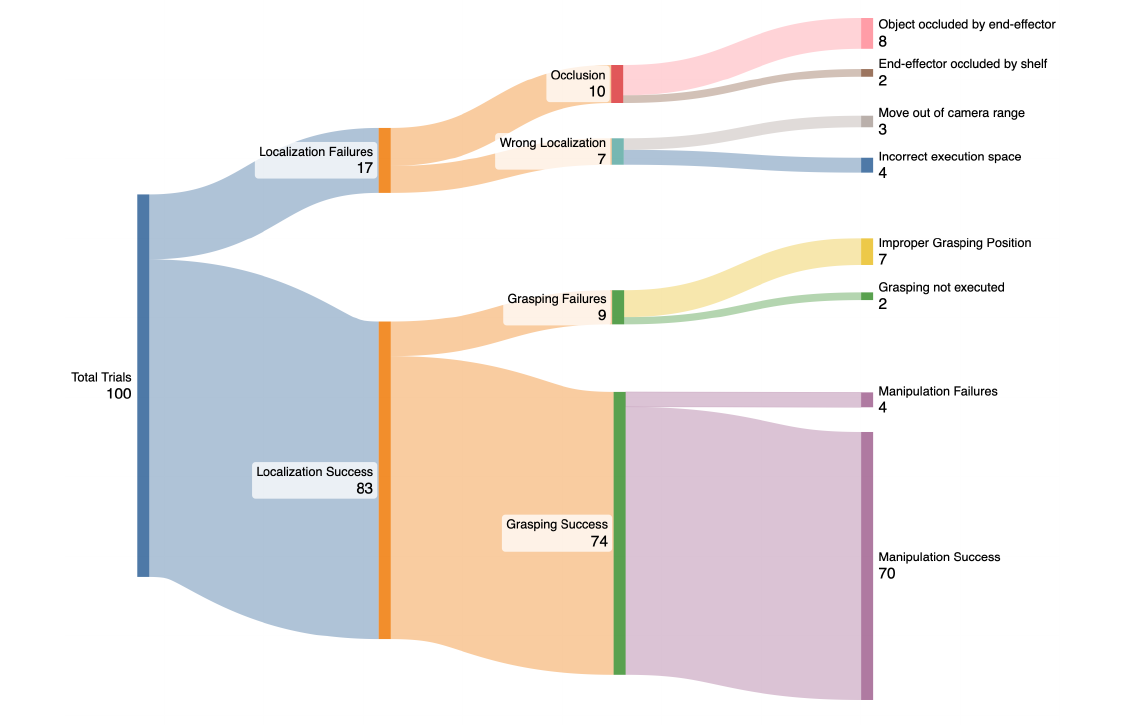}
    \caption{\textbf{Detailed failure case analysis on real-world robot experiments.}}
    \label{fig:sankey}
\end{figure}

During the 20 testing runs of the five tasks in the kitchen environment, we record the causes behind each failure. Subsequently, we present these failure causes in a Sankey diagram, exemplified in \cref{fig:sankey}. 
Among our 100 experimental trials, 17 failure cases are attributed to localization errors. Among them, 7 failure cases are caused by incorrect localization, while the remaining 10 are due to occlusion, including instances where objects are occluded by the end-effector and the end-effector is occluded by the shelf. 
We employ a third-person perspective camera configuration, which is prone to occlusion in the captured images.
Dealing with occlusion issues in robot manipulation scenarios poses a significant challenge. In the trials of localization success, there are also 9 instances of grasping failures, primarily due to incorrect grasping position.

\paragraph{Generalization on position and lighting condition.}
To further validate the generalization ability of our model, we further benchmark MPI with other approaches under object position and lighting condition variation. 
The experimental setting is depicted in~\cref{fig:generaliation_setting}. Evaluations are conducted at ten marked locations to ensure consistency and fairness. 

\begin{figure}[t]
    \centering
    \includegraphics[width=0.56\linewidth]{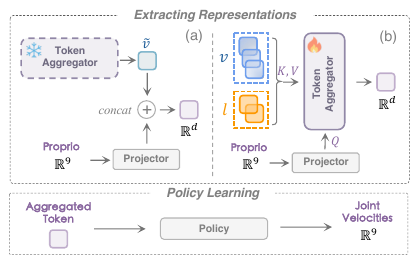}
    \caption{\textbf{Representation extraction and policy learning pipeline for downstream adaptation on visuomotor control tasks.} We broadly categorize extraction approaches into: \textbf{a) Pre-aggregation}, where we incorporate the aggregator within the encoder itself without further training for adaptation, and \textbf{b) Post-aggregation}, involving the learning of an aggregator on downstream data to extract from full-length embeddings.} 
    \label{fig:adaptation}
\end{figure}

\begin{table}[t]
  \small % \footnotesize \small \normalsize
  \centering
  \caption{\textbf{The average success rate on Franka Kitchen with different feature aggregation strategies.}
  Our proposed Proprio-conditioned MAP brings notable improvement for all ViT-based models on visuomotor control tasks.
  }
  \label{table:ablation_AggregationStrategy}
  \setlength{\tabcolsep}{6mm}{
  \begin{tabular}{l|c|cc}
    \toprule 
    \multirow{2}{*}{Method}& \multirow{2}{*}{Backbone} & \multicolumn{2}{c}{Aggregation Strategy} \\
    % \cmidrule{3-4}
    & & Original & Ours \\
    \midrule
    R3M~\cite{nair2023r3m}  &ResNet50  &69.5 & 66.7 ($\downarrow$2.8)\\
    MVP~\cite{radosavovic2023mvp}  &ViT-Base  &42.1 & 71.7 ($\uparrow$29.6)\\
    Voltron~\cite{karamcheti2023voltron}  &ViT-Base &31.8 & 70.5 ($\uparrow$38.7)\\
    EVA-02~\cite{fang2023eva02} &ViT-Base &46.0 & 71.6 ($\uparrow$25.6)\\
    \midrule
    \baseline{\modelname (Ours)}  & \baseline{ViT-Small} & \baseline{70.6} & \baseline{76.5 ($\uparrow$5.9)} \\
    \baseline{\modelname (Ours)}  & \baseline{ViT-Base} & \baseline{\textbf{72.9}} & \baseline{\textbf{78.9} ($\uparrow$6.0)} \\
    \bottomrule
  \end{tabular}}
\end{table}
\begin{table}[t]
    \centering
    \small
    \setlength{\tabcolsep}{7mm}{
    \caption{\textbf{The representation extraction strategy adopted by different methods.} In our experiments, we obtain full-length visual embeddings by capturing features before Average Pooling for R3M using a ResNet50 backbone, and sequence outputs for all methods employing a ViT backbone.}
    \label{supp:strategy}
    \begin{tabular}{l|c | c}
    \toprule
    Method & Backbone & Representation Extraction Strategy \\
    \midrule
         R3M~\cite{nair2023r3m} & ResNet50 & Average Pooling 2D\\
         MVP~\cite{radosavovic2023mvp} & ViT-Small &\texttt{[cls]} Token \\
         Voltron~\cite{karamcheti2023voltron} & ViT-Small & Average Pooling 1D \\
    \midrule
         \baseline{MPI (Ours)} & \baseline{ViT-Small} & \baseline{Token Aggregator} \\
    \bottomrule
    \end{tabular}}
\end{table}

\subsection{In-Depth Analysis into Franka Kitchen}
\label{sec:details_franka_kitchen}

%\subsubsection{Extracting Representations}

How to properly apply the learned representation is also important. Especially for Vision Transformers (ViTs) whose output is a sequence of embeddings representing each divided region of the image, it is essential to aggregate the sequence into one unified representation compatible with multi-layer perceptron (MLP) policy networks. Drawing inspiration from Voltron~\cite{karamcheti2023voltron}, we also employ multi-headed attention pooling (MAP)~\cite{lee2019set} as the base structure to aggregate the sequence of tokens. Taking a step further, we integrate this aggregation methodology into the pre-training stage, as elucidated in \cref{sec:method_mpi}. This integration facilitates zero-shot transferability to downstream tasks, obviating the necessity for additional parameter learning of aggregator~(\cref{fig:adaptation}(a) in the top left). The utilization of the aggregated visual token, denoted as $\Tilde{v}$, not only enhances the adaptability of the model but also enables equitable comparisons and analyses with other models and methods that produce only one feature vector.

The features required for policy learning vary across distinct stages of an interaction process.
To achieve adaptive feature extraction based on ego state, we further propose the \textbf{Proprio-conditioned MAP} (\cref{fig:adaptation}(b) in top right) where the latent query will first be initialized by projected proprioception signal, instead of random initialization as employed in the original MAP. Furthermore, the derived feature can be used directly for imitation, obviating the need for concatenation of visual representation and proprioception signal as the input for policy networks, which was previously customary in practice. As demonstrated in~\Cref{table:ablation_AggregationStrategy}, empirical investigations conducted across various models consistently reveal its superior performance compared to alternative aggregation strategies such as the \texttt{[cls]} token~\cite{radosavovic2023mvp}, global average pooling~\cite{nair2023r3m}, and the original MAP~\cite{karamcheti2023voltron}.

\begin{table}[H]
  \small
  \centering
  \caption{\textbf{The average precision~(\%) under three IoU thresholds on Referring Expression Grounding task.} We report here only the results of smaller variants~(ResNet50 for R3M, and ViT-Small for MVP, Voltron and ours). *: We leverage aggregated visual embedding from the encoder. MPI achieves the best results using both aggregated and full visual features.}
  \label{table:supp_refer}
  \setlength{\tabcolsep}{5mm}{
  \begin{tabular}{l| l | c c c}
    \toprule
         \multicolumn{1}{l|}{\multirow{2}{*}{Method}} & \multirow{2}{*}{Embedding} &  \multicolumn{3}{c}{Average Precision~(AP)} \\
         % \cmidrule{3-5}
         & & 0.25 IoU & 0.5 IoU &0.75 IoU \\
         \midrule
        % R3M& $\mathbb{R}^{2048}$ & 44.35 & 23.19 & 6.28\\
        
        % MVP & $\mathbb{R}^{384}$ & 93.07 &85.32 &60.37\\
        R3M~\cite{nair2023r3m}* & $\mathbb{R}^{2048}$ & 85.27 & 71.79 & 42.66\\
        R3M~\cite{nair2023r3m} &$\mathbb{R}^{49 \times 2048}$ & 90.35 & 81.13 & 55.56 \\

        Voltron~\cite{karamcheti2023voltron}*& $\mathbb{R}^{384}$ & 90.97& 80.83& 53.61\\
        Voltron~\cite{karamcheti2023voltron} &$\mathbb{R}^{196 \times 384}$ & 92.93& 84.70& 57.61\\

        MVP~\cite{radosavovic2023mvp}* & $\mathbb{R}^{384}$ & 93.07 &85.32 &60.37\\
        MVP~\cite{radosavovic2023mvp} &$ \mathbb{R}^{197 \times 384}$ & 94.43 & 88.98 &67.52 \\
        \midrule
        \baseline{MPI (Ours)*}  & \baseline{$\mathbb{R}^{384}$} &\baseline{\bf{96.29}} &\baseline{\bf{92.10}} &\baseline{71.87}\\
        \baseline{MPI (Ours)} & \baseline{$\mathbb{R}^{196 \times 384}$} & \baseline{96.04}& \baseline{92.05} & \baseline{\bf{74.40}} \\
    \bottomrule
    \end{tabular}}
\end{table}

\noindent
We hope that this improvement will aid future research and provide valuable insights.

\subsection{A Closer Look to Referring Expression Grounding}
\label{sec:details_reg}

As discussed in~\cref{sec:details_franka_kitchen}, the representation extraction strategy is worth studying for downstream adaptations. In our paper, we primarily focus on the default feature extraction and usage of previous methods~\cite{nair2023r3m,radosavovic2023mvp,karamcheti2023voltron}. Additionally, we explore the effectiveness of incorporating aggregated features directly into downstream tasks and utilizing full-length visual embedding to complement our analysis. The extraction strategies leveraged by different models are given in~\Cref{supp:strategy}. Following the approach of Voltron, we utilize the MAP block, as described by~\citet{lee2019set}, to process the concatenation of visual embedding (specified dimension in~\Cref{table:supp_refer}) and language embedding produced by DistillBERT~\cite{devlin2018bert}. This is followed by an MLP for bounding box regression. We leverage the officially provided model weights for all baselines. For training details, we train all models for 10 epochs using AdamW as the optimizer, with an initial learning rate fixed at 0.001.

The complete results are in~\Cref{table:supp_refer}. It can be observed that our method yields superior outcomes under three IoUs by leveraging both post-aggregation and full-length visual embedding. It is worth noting that our method sustains high performance even when incorporating aggregated features, in contrast to other methods that typically exhibit performance declines. This underscores the efficacy of our proposed token aggregator module. Despite the additional parameters introduced~(about 0.5M for ViT-Small), the token aggregator is training-free for downstream adaptations and can be zero-shot transferred to various downstream tasks.

% \subsection{Visualization}
% \label{sec:visualization}

\subsection{More Related Works}
\label{sec:more_related_works}
\paragraph{Vision-language models in Robotics.}
RT-1~\cite{brohan2022rt1} collects 130k robot demonstrations, with a fleet of 13 robots over 17 months, and establishes a transformer-based end-to-end pipeline to output control signals from visual input and language instructions. 
PaLM-E~\cite{driess2023palm_e} utilizes an embodied multi-modal model with 562B parameters to generate high-level planning instruction. 
It proves that co-fine-tuning with a mixture of in-domain robotic scene VQA data and web-based VQA data enhances the zero-shot generalization capability.
Inspired by PaLM-E, RT-2~\cite{brohan2023rt2} co-fine-tunes the vision-language model with both the robot trajectory data and the web data. RT-2 goes a step further by expressing robot actions as text tokens and seamlessly integrates them into the training pipeline of vision-language models.
The aforementioned studies require a large amount of both web-scale vision-language pairs and low-level robot actions. Such solutions present challenges for robotics practitioners seeking to adopt them.
RoboFlamingo~\cite{li2023roboflamingo} only requires a small number of manipulation demonstrations for fine-tuning.
There will be a surge of research endeavors focused on data-efficient vision-language models in the field of robotics.
All of these vision-language models rely on the ViT encoder to extract distinct visual cues, the ViT encoder obtained via our proposed pre-train method has the potential to enhance visual language models with interaction-oriented features. 

\clearpage
\newpage
% \bibliographystyle{plainnat}
% \bibliography{short, references}

\end{document}